\pdfoutput=1
\documentclass{article}
\PassOptionsToPackage{numbers}{natbib}

\usepackage[preprint]{neurips_2026}
\usepackage{xcolor}
\usepackage{colortbl}
\usepackage{algorithm}
\usepackage{algpseudocode}
\usepackage{amssymb}
\usepackage{amsmath}
\usepackage{amsfonts}
\usepackage{nicefrac}
\usepackage{enumitem}
\usepackage{tcolorbox}
\tcbuselibrary{listings,breakable,skins}

\usepackage{listings}
\usepackage{multirow}
\usepackage{booktabs}
\usepackage{wrapfig}

\usepackage[normalem]{ulem}
\useunder{\uline}{\ul}{}
\usepackage{subcaption}
\usepackage{caption}
\usepackage{bm}
\usepackage[british, american]{babel}

\usepackage{tikz}
\usetikzlibrary{shapes.geometric}

\usepackage{pifont}

\definecolor{mygreen}{RGB}{0,150,0}
\definecolor{myred}{RGB}{200,0,0}

\newcommand{\cmark}{\textcolor{mygreen}{\ding{51}}}
\newcommand{\xmark}{\textcolor{myred}{\ding{55}}}

\definecolor{citecolor}{HTML}{0071BC}
\definecolor{linkcolor}{HTML}{ED1C24}
\definecolor{linkpink}{RGB}{210, 80, 140}

\usepackage[pagebackref=true,breaklinks=true,colorlinks,bookmarks=false,citecolor=citecolor,linkcolor=linkcolor,urlcolor=gray]{hyperref}

\usepackage{arydshln}

\usepackage[utf8]{inputenc}
\usepackage[T1]{fontenc}
\usepackage{url}
\usepackage{graphicx}
\usepackage{microtype}
\usepackage[capitalize]{cleveref}
\crefname{section}{Sec.}{Secs.}
\Crefname{section}{Sec.}{Secs.}
\crefname{subsection}{Sec.}{Secs.}
\Crefname{subsection}{Sec.}{Secs.}
\crefname{subsubsection}{Sec.}{Secs.}
\Crefname{subsubsection}{Sec.}{Secs.}
\crefname{figure}{Fig.}{Figs.}
\Crefname{figure}{Fig.}{Figs.}
\crefname{table}{Tab.}{Tabs.}
\Crefname{table}{Tab.}{Tabs.}
\crefname{equation}{Eq.}{Eqs.}
\Crefname{equation}{Eq.}{Eqs.}
\crefname{algorithm}{Alg.}{Algs.}
\Crefname{algorithm}{Alg.}{Algs.}

\definecolor{darkteal}{RGB}{0, 100, 80}
\definecolor{lightgraytext}{RGB}{180, 180, 180}
\definecolor{citegray}{RGB}{100, 100, 100}

\title{RankE: End-to-End Post-Training for Discrete Text-to-Image Generation with Decoder Co-Evolution}

\author{
\begin{tabular}{c}
\textbf{Siyong Jian\textsuperscript{$\star$}$^{1}$ \quad Siyuan Li\textsuperscript{$\star$}$^{1,2}$ \quad Luyuan Zhang\textsuperscript{$\star$}$^{3}$ \quad Zedong Wang$^{4}$} \\[0.3em]
\textbf{Xin Jin$^{1}$ \quad Ying Li$^{1}$ \quad Cheng Tan\textsuperscript{$\dagger$}$^{5}$ \quad Huan Wang\textsuperscript{$\dagger$}$^{1}$} \\[0.5em]
\normalfont
$^1$Westlake University \quad
$^2$Zhejiang University \quad
$^3$Tsinghua University \\[0.1em]
\normalfont
$^4$Hong Kong University of Science and Technology \quad
$^5$Shanghai AI Lab \\[0.3em]
\normalfont \textsuperscript{$\star$}Equal contribution. \quad \textsuperscript{$\dagger$}Corresponding author. \\[0.3em]
{\hypersetup{urlcolor=linkpink}
Project page: \url{https://syjmelody.github.io/RankE/}}
\end{tabular}
}

\begin{document}

\maketitle

\begin{abstract}
Discrete autoregressive (AR) text-to-image (T2I) models pair a VQ tokenizer with an AR policy, and current post-training pipelines optimize only the policy while keeping the VQ decoder frozen. %
Recent diffusion T2I work, exemplified by REPA-E, has shown that the VAE itself constitutes a key alignment bottleneck, yet no analogous investigation exists for discrete AR models. %
We show that policy-only optimization induces \textit{Latent Covariate Shift}: as the policy evolves, the resulting token distribution diverges from the ground-truth distribution on which the decoder was trained, such that reward scores improve while decoded image quality degrades. %
To address this mismatch, we propose RankE, the first end-to-end post-training framework for discrete T2I generation. %
Rather than optimizing the policy against a fixed decoder, RankE co-evolves both components through alternating optimization: each module maximizes a ranking-based alignment objective while being regularized by a stability-preserving anchor suited to its parameter space. %
This co-evolution breaks the fidelity--alignment trade-off that plagues frozen-decoder approaches: on LlamaGen-XL (775M), standard RL improves CLIP but degrades FID, whereas RankE improves both simultaneously (FID 15.21, CLIP 33.76 on MS-COCO 30K). %
Consistent gains on Janus-Pro (1B) confirm that decoder co-evolution reliably converts reward optimization into pixel-space quality improvements. %
\end{abstract}

\section{Introduction}
\label{sec:intro}

Discrete autoregressive (AR) text-to-image (T2I) models factorize image generation into two stages: a VQ tokenizer~\citep{van2017neural,esser2021taming} maps images to discrete codebook entries, and an AR policy models the resulting token sequences via next-token prediction~\citep{kaplan2020scaling,sun2024llamagen}. %
This formulation enables unified multimodal architectures~\citep{team2024chameleon,deepseek2025januspro} and directly inherits the favorable scaling behavior and infrastructure of large language models. %
The alignment of these models increasingly relies on post-training~\citep{simplear2024,zhang2026group,jiang2025t2i}, which conventionally optimizes only the AR policy while keeping the VQ decoder frozen. %

\begin{figure}[h]
    \centering
    \includegraphics[width=0.95\linewidth]{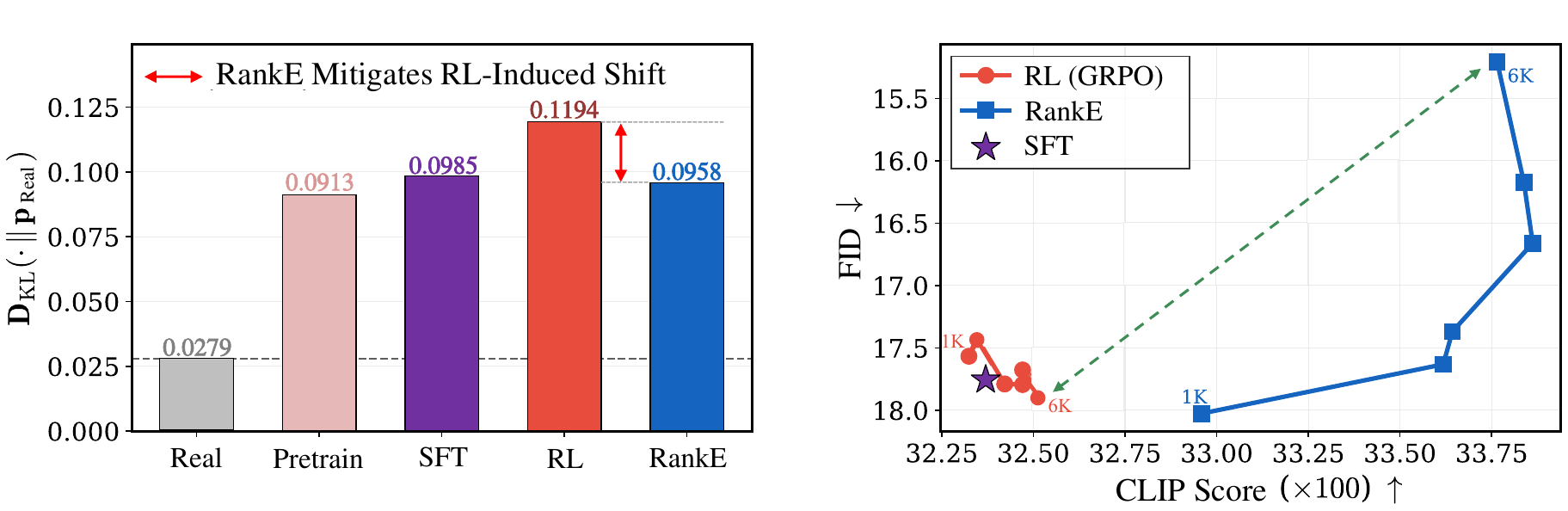}
\caption{%
\textbf{Latent Covariate Shift intensifies under RL and is mitigated by RankE.} %
\textbf{Left:} KL divergence between each model's VQ token distribution and that of $5{,}000$ real MS-COCO images; %
the dashed line marks \textit{Real}, a natural-variation lower bound computed from two independently sampled real-image sets encoded by the same frozen tokenizer. %
The shift grows progressively across pre-training, SFT, and RL---standard RL widens the endpoint gap by $21\%$ over SFT---while RankE returns it to the SFT level via decoder co-evolution; per-step dynamics are shown in Fig.~\ref{fig:analyse_combine_shift_entropy}. %
\textbf{Right:} CLIP--FID trajectory across training checkpoints. %
Standard RL improves CLIP at the cost of stagnating FID, exposing a systematic fidelity--alignment tension caused by the frozen decoder; RankE breaks this tension, simultaneously raising CLIP and lowering FID throughout training. %
}
    \label{fig:combined}
    \vspace{-1.5em}
\end{figure}

This frozen-decoder convention is increasingly out of step with recent progress on the continuous T2I side. %
Diffusion methods such as REPA-E~\citep{leng2025repa} have begun to unlock the VAE for joint optimization with the denoiser, thereby lifting a frozen-decoder assumption that has long been treated as a default in latent generative modeling. %
In discrete AR, the picture is precisely the opposite: existing post-training methods~\citep{jiang2025t2i,simplear2024,zhang2026group,liao2025va} universally freeze the VQ decoder and optimize only the AR policy. %

\begin{figure}[t]
    \centering
    \includegraphics[width=\textwidth]{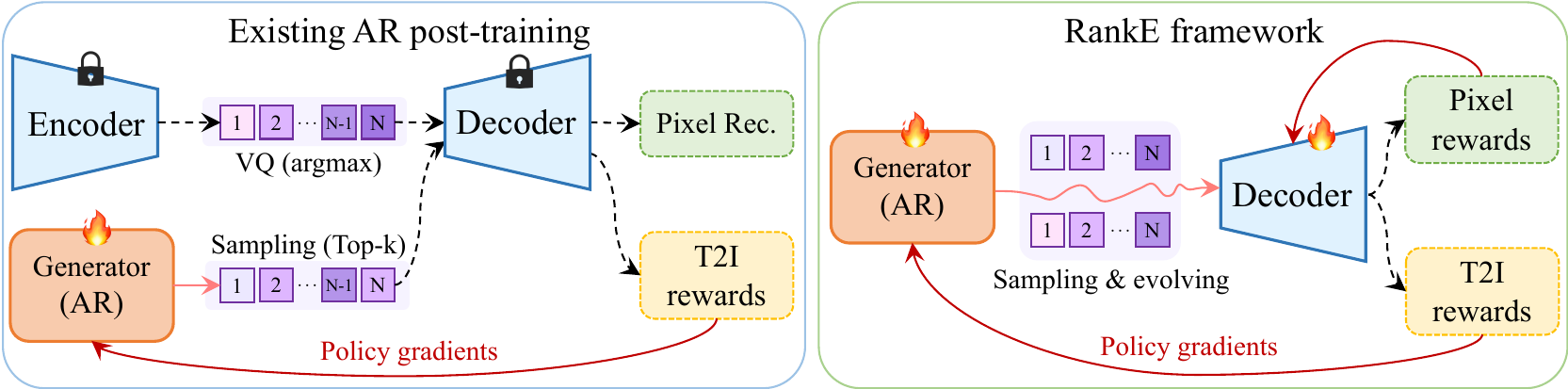}
    \caption{
    Comparison of existing AR post-training and RankE framework. \textbf{Left:} After pre-training the VQ tokenizer, the existing works \citep{deepseek2025januspro, simplear2024} only optimize the AR generator with the sampled latent rollouts by the RL ranking loss (policy gradients) with the T2I rewards while freezing the decoder, which will cause the Latent Covariate Shift between the generator and the decoder.
    \textbf{Right:} RankE applies an alternative training pipeline to co-evolve both the generator and the decoder by RL objectives with the T2I rewards (for the generator \& decoder) and pixel rewards (for the decoder). By unlocking the decoder, this joint optimization directly mitigates this shift, ensuring that the latent representations remain firmly grounded in high-fidelity visual outputs.
    }
    \label{fig:intro_teaser}
    \vspace{-4mm}
\end{figure}

We identify the underlying mismatch as \textit{Latent Covariate Shift} (\Cref{fig:intro_teaser}~(a)). %
During tokenizer pre-training, the VQ decoder is trained exclusively on deterministic ground-truth codes $z_{\mathrm{gt}}=\mathrm{Quantize}(E(x))$~\citep{van2017neural,esser2021taming}, which occupy a restricted, low-variance region of the latent space~\citep{razavi2019generating}. %
At inference, however, the same decoder receives tokens sampled from the AR policy, $\hat{z}\sim\pi_\theta(\cdot\mid y)$, whose distribution progressively diverges from this regime as the policy evolves under reward pressure. %
This divergence produces a fidelity--alignment trade-off that policy-side tuning alone cannot resolve: GRPO~\citep{shao2024deepseekmath} applied to LlamaGen-XL~\citep{sun2024llamagen} improves CLIP yet degrades FID across checkpoints (\Cref{fig:combined}, right), and the KL divergence against ground-truth token statistics (\Cref{fig:combined}, left) confirms that standard RL substantially widens the distributional gap relative to SFT. %
Unlike exposure bias~\citep{bengio2015scheduled,ranzato2015sequence}, which concerns the input context of the generator, Latent Covariate Shift targets the input distribution of the decoder---a mismatch that no amount of policy-level tuning can resolve. %

Resolving this shift requires updating the decoder jointly with the policy, but direct end-to-end optimization is blocked by two non-differentiable operations along the generation chain: categorical sampling at $z\sim\pi_\theta$ and VQ quantization~\citep{jang2016categorical,bengio2013estimating}. %
Together, these operations sever the gradient path from pixel-space rewards to policy parameters (\Cref{fig:intro_teaser}~(a))---a barrier that simply does not arise in continuous diffusion models, where the generation chain remains fully differentiable~\citep{clark2024directly,prabhudesai2023aligning,leng2025repa}. %
Standard surrogates~\citep{bengio2013estimating,huh2023straightening,jang2016categorical} introduce non-trivial gradient bias or training instability at the codebook scales used by modern visual tokenizers~\citep{kaiser2018fast}. %
Consequently, all existing post-training methods for discrete AR resort to a frozen decoder and inherit the resulting cost in fidelity. %

We introduce \textbf{RankE} (\textbf{Rank}ing-based \textbf{E}nd-to-end alignment), the first end-to-end post-training framework for discrete AR T2I models that jointly evolves the policy and the decoder without differentiating through the discrete bottleneck. %
The name reflects two ranking-based mechanisms that operate at complementary granularities: a \emph{token-level} ranking objective (group-relative advantages in GRPO) drives the policy update, and a \emph{pixel-level} ranking objective (a reward-weighted adversarial loss, \emph{Rank-GAN}) drives the decoder update. %
As illustrated in \Cref{fig:intro_teaser}~(b), RankE employs an alternating optimization strategy that admits a Generalized EM interpretation (\Cref{subsec:coevolution}). %
In the \emph{policy stage}, the AR generator is updated via group-relative preference optimization~\citep{shao2024deepseekmath} with KL regularization. %
In the \emph{decoder stage}, the VQ decoder is adapted on policy-sampled latents through Rank-GAN and EMA-anchored consistency regularization, which together prevent the decoder from drifting away from its reconstruction prior. %

By allowing the decoder to continuously track the evolving token distribution of the policy, RankE absorbs Latent Covariate Shift during training and breaks the fidelity--alignment trade-off (\Cref{fig:combined}, right): on LlamaGen-XL (775M), RankE simultaneously improves FID to $15.21$ and CLIP to $33.76$ on MS-COCO 30K, whereas standard RL improves alignment at the expense of fidelity. %
On Janus-Pro-1B and under the HPSv2 reward, RankE consistently improves alignment (CLIP/HPSv2) and zero-shot GenEval over the standard-RL baseline, further confirming the generality of the approach. %

Our contributions are summarized as follows:
\begin{itemize}[leftmargin=2.0em, nosep, itemsep=3pt]
    \item We identify \textit{Latent Covariate Shift}---a decoder-side distribution mismatch distinct from generator-side exposure bias---and demonstrate that RL post-training exacerbates this shift.

    \item We propose RankE, the first end-to-end post-training framework for discrete AR T2I models. RankE co-evolves the AR policy and the VQ decoder via alternating optimization, enabling reward signals to propagate through the discrete token--pixel interface.

    \item We demonstrate that RankE simultaneously improves fidelity \emph{and} alignment across two model backbones (LlamaGen-XL, Janus-Pro), three evaluation dimensions (FID, CLIP/HPSv2, GenEval), and two reward functions (CLIP, HPSv2), consistently breaking the fidelity--alignment trade-off observed in frozen-decoder baselines.
\end{itemize}

\section{Related Work}
\label{sec:related_work}

\paragraph{Post-Training and Alignment in T2I}
Post-training for T2I has matured rapidly in the diffusion family. %
Online RL~\citep{black2024training,fan2023dpok}, offline preference optimization~\citep{wallace2023diffusion}, and direct reward fine-tuning~\citep{xu2024imagereward,clark2024directly,prabhudesai2023aligning} all exploit a key structural property that is unavailable in discrete AR: %
the denoising chain is differentiable end-to-end, so reward gradients can flow from pixel space back to the generator via $\nabla_\theta R(\mathbf{x})$. %
More recently, REPA-E~\citep{leng2025repa} goes one step further by unlocking the VAE for joint optimization with the denoiser, lifting the frozen-decoder assumption that has long been treated as a default in latent generative modeling, and demonstrating that decoder adaptation yields gains in both fidelity and alignment. %
For discrete AR, by contrast, post-training is far less developed. %
Methods such as T2I-R1~\citep{jiang2025t2i}, SimpleAR~\citep{simplear2024}, GCPO~\citep{zhang2026group}, and VA-$\pi$~\citep{liao2025va} apply GRPO~\citep{shao2024deepseekmath} to the AR policy. %
Without exception, these methods keep the VQ decoder frozen: reward is computed in pixel space, yet only the policy is updated. %
A complementary line of work~\citep{levine2018reinforcement,korbak2022rl} reframes KL-regularized RL as variational inference under a reward-induced log-likelihood; we leverage this perspective to ground the alternation of RankE as a Generalized EM procedure (\S\ref{subsec:coevolution}). %
As shown in \S\ref{sec:method}, this frozen-decoder regime is precisely where Latent Covariate Shift is most severe and the FID/CLIP trade-off most pronounced. %
A broader survey of discrete visual tokenizers, AR generator factorizations, and the gradient barrier is deferred to Appendix~\ref{app:related_extended}. %

\section{Method}
\label{sec:method}

\subsection{Problem Formulation}
\label{subsec:formulation}

A discrete AR T2I system consists of an autoregressive token policy $\pi_\theta(z\mid y)$ and a VQ decoder $D_\phi$ that renders codes into images via $x = D_\phi(z)$. %
Given a reward function $r$ that scores text--image alignment or human preference~\citep{xu2024imagereward,wu2023hpsv2}, we seek to maximize
\begin{equation}
    \max_{\theta,\phi}\;
    \mathbb{E}_{y\sim\mathcal{D},\,z\sim\pi_\theta(\cdot|y)}
    \!\left[\,r\!\left(D_\phi(z),\,y\right)\,\right].
    \label{eq:e2e_obj}
\end{equation}
This single objective ties both modules to one pixel-space reward, yet categorical sampling at $z\sim\pi_\theta$ and VQ quantization~\citep{jang2016categorical,bengio2013estimating} sever the gradient path: signals flow into the decoder but cannot reach the policy. %
Rather than forcing this bottleneck with a biased surrogate, RankE alternates \emph{around} it---each module is updated with the signal natural to its own parameter space, and reward information crosses the gap through the interleaving of the two updates. %
\Cref{subsec:coevolution} casts this alternation as a unified regularized objective and connects it to a Generalized EM procedure, and the decoder design. %

\subsection{Alternating Co-Evolution Around the Discrete Bottleneck}
\label{subsec:coevolution}

\paragraph{A unified two-stage objective.}
Although the two updates live on incompatible parameter spaces---discrete tokens for $\pi_\theta$ and continuous pixels for $D_\phi$---they share a common structure. %
The updated parameter $\Psi\in\{\theta,\phi\}$ maximizes a regularized alignment objective
\begin{equation}
    \max_{\Psi\in\{\theta,\phi\}}\;
    \mathcal{J}(\Psi) =
    \underbrace{\mathbb{E}\!\left[\,\mathcal{A}_{\Psi}\,\right]}_{\text{ranking-based alignment}}
    \;-\;
    \underbrace{\lambda\,\Omega(\Psi)}_{\text{stability-preserving regularization}},
    \label{eq:unified}
\end{equation}
where $\mathcal{A}_\Psi$ pushes $\Psi$ toward the reward-favored region and $\Omega(\Psi)$ keeps it tethered to a trusted prior. %
Crucially, $\mathcal{A}_\Psi$ is implemented through \emph{relative ranking} rather than absolute reward magnitude: at every step, we draw $G$ rollouts from the same prompt, score them with $r$, and update $\Psi$ in the direction of higher-ranked samples. %
Stage~1 applies this principle at the \emph{token} level on $\theta$, and Stage~2 applies it at the \emph{pixel} level on $\phi$. %
This shared ranking principle---per-prompt comparison of $G$ rollouts at two complementary granularities---is what the name \emph{RankE} encodes. %
The two stages run alternately within each round and across $K$ rounds, so reward information crosses the discrete bottleneck through this alternation rather than through any single gradient path. %

\paragraph{Stage~1: token-level ranking via GRPO.}
With the decoder fixed, we update the policy using Group Relative Policy Optimization~\citep{shao2024deepseekmath}, which converts reward scalars into a per-prompt ranking at the token level. %
For each prompt $y$, we draw $G$ rollouts $\{z_i\}_{i=1}^G\sim\pi_\theta(\cdot\mid y)$, decode them with the frozen $D_\phi$, score them under $r$, and form group-normalized advantages $A_i=(r_i-\mu_r)/\sigma_r$~\citep{williams1992simple}. %
The advantage $A_i$ itself constitutes a ranking signal: its sign records whether rollout $i$ beats or trails its peers under the same prompt, and its magnitude records by how much. %
The resulting loss
\begin{equation}
\mathcal{L}_{\pi}(\theta) =
-\,\mathbb{E}_{y}\!\left[\,
\underbrace{
\frac{1}{G}\sum_{i=1}^{G}
\min\!\big(\rho_i A_i,\;\mathrm{clip}(\rho_i,1{\pm}\epsilon)\,A_i\big)
}_{\mathcal{A}_\theta:\;\text{token-level ranking}}
\;-\;
\underbrace{
\beta\,\mathbb{D}_{\mathrm{KL}}\!\big(\pi_\theta\,\|\,\pi_{\mathrm{ref}}\big)
}_{\Omega_\theta:\;\text{stability anchor}}
\,\right],
\label{eq:policy}
\end{equation}
maps cleanly onto Eq.~\ref{eq:unified}: the clipped advantage term is the token-level ranking signal $\mathcal{A}_\theta$, and the KL against an EMA reference~\citep{gao2023scaling,coste2023reward} serves as the stability anchor $\Omega_\theta$. %
Here, $\rho_i=\pi_\theta(z_i\mid y)/\pi_{\theta_{\mathrm{old}}}(z_i\mid y)$ denotes the PPO importance ratio~\citep{schulman2017proximal}. %

\begin{figure}[!t]
    \centering
    \includegraphics[width=1.0\textwidth]{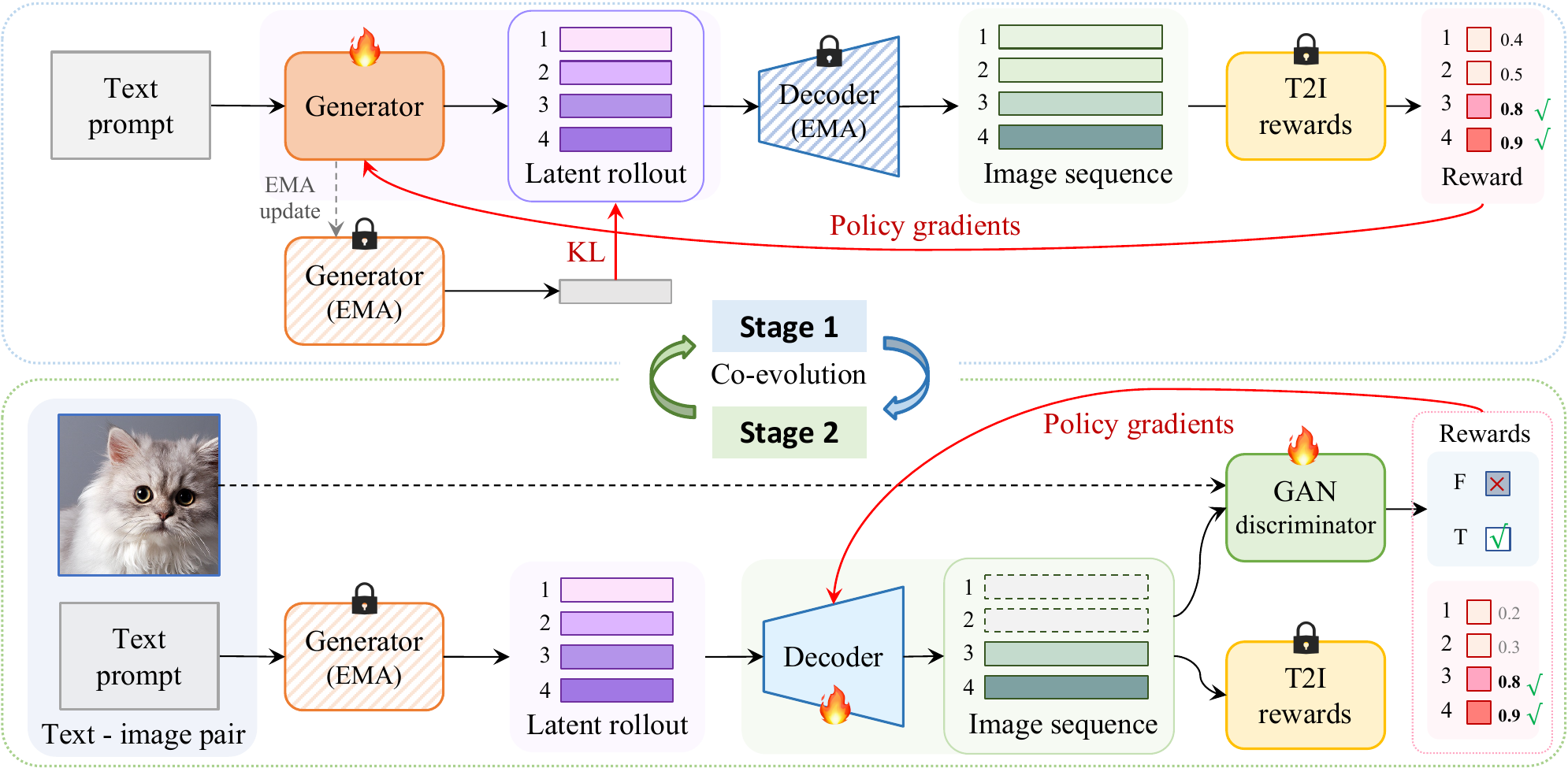}
    \caption{%
    Overview of the RankE co-evolution framework.
    \textbf{Stage~1 (Policy Alignment):}
    The AR policy is updated via policy-gradient optimization to maximize a human-preference reward, with KL regularization toward an EMA reference policy.
    The VQ decoder is frozen; this stage improves latent-space alignment but widens the mismatch with the decoder training distribution.
    \textbf{Stage~2 (Decoder Adaptation):}
    The policy is frozen and the VQ decoder is updated on policy-sampled latents via reward-weighted adversarial supervision (Rank-GAN), differentiable reward gradients, and EMA consistency regularization.
    Alternating the two stages couples policy improvement with decoder adaptation, absorbing Latent Covariate Shift at each round.}
    \label{fig:method_teaser}
\end{figure}

\paragraph{Stage~2: pixel-level ranking, at a glance.}
With the policy fixed, we allow the decoder to track its evolving token distribution. %
The same $G$ rollouts that Stage~1 has just ranked in \emph{token} space are now re-ranked in \emph{pixel} space: decoded images preferred by the reward model receive a larger weight in the decoder update, less-preferred samples are down-weighted, and the gradient pulls $D_\phi$ toward outputs resembling the top-ranked decodings. %
Mirroring the structure of Stage~1, the decoder loss decomposes into a ranking-based alignment block and a manifold-anchored regularization block:
\begin{equation}
\mathcal{L}_{D}(\phi) =
\underbrace{
\lambda_d\,\mathcal{L}_{\mathrm{reward}}
\;+\;\lambda_g\,\mathcal{L}_{\mathrm{Rank\text{-}GAN}}
}_{\mathcal{A}_\phi:\;\text{pixel-level ranking}}
\;+\;
\underbrace{
\lambda_r\,\mathcal{L}_{\mathrm{recon}}
\;+\;\lambda_c\,\mathcal{L}_{\mathrm{consist}}
}_{\Omega_\phi:\;\text{manifold anchor}}.
\label{eq:decoder}
\end{equation}
At this level, the symmetry with Stage~1 is exact: $\mathcal{A}_\phi$ ranks policy-sampled decodings in pixel space via a reward-weighted adversarial signal (\emph{Rank-GAN}), while $\Omega_\phi$ anchors the decoder to the deterministic ground-truth manifold on which the tokenizer was trained. %
The concrete instantiation of the pixel-level ranking, together with the role of each loss term, is the subject of \Cref{subsec:decoder_design}. %

\paragraph{Why alternation, and a GEM view.}
A natural alternative is to fuse $\mathcal{L}_\pi$ and $\mathcal{L}_D$ into a single gradient step. %
We avoid this because $\nabla_\theta$ is a high-variance policy-gradient estimator~\citep{williams1992simple}, whereas $\nabla_\phi$ is a low-variance differentiable signal; mixing them in one step couples their effective step sizes in ways that no learning-rate schedule can disentangle. %
The deeper reason is principled: the alternation realizes a Generalized EM procedure~\citep{levine2018reinforcement,korbak2022rl}, in which Stage~1 acts as a variational E-step on $\pi_\theta$ under a reward-induced log-likelihood, and Stage~2 acts as a MAP M-step on $\phi$ with $\mathcal{L}_{\mathrm{recon}}$ and $\mathcal{L}_{\mathrm{consist}}$ serving as the log-prior on the decoder manifold. %
Under this view, RankE inherits standard GEM convergence guarantees~\citep{neal1998view,wu1983convergence}, whereas a fused joint update would forfeit them. %
Algorithm~\ref{alg:joint_training} summarizes the alternating schedule, and the formal derivation is provided in Appendix~\ref{app:em}. %

\subsection{Decoder Adaptation: Reward-Driven Alignment Meets Manifold Anchoring}
\label{subsec:decoder_design}

The decoder is where Latent Covariate Shift is actually absorbed, and where the design choices that distinguish RankE from a frozen-decoder baseline reside. %
The two blocks of Eq.~\ref{eq:decoder} mirror the two blocks of GRPO, but each is internally richer; we unpack them in turn. %

\subsubsection{Reward-driven alignment ($\mathcal{A}_\phi$)}
\label{subsubsec:decoder_alignment}

The alignment block plays the same role for the decoder as the group-relative advantage plays for the policy: it injects reward information into the parameter update. %
Because T2I rewards come in two flavors---differentiable scorers such as CLIP and black-box scorers such as HPSv2---we adopt two complementary channels rather than one, with the choice dictated by the reward. %

\textbf{Differentiable channel: direct reward back-propagation.}
When the reward $R$ admits gradients, the decoder offers a fully differentiable path from latents to scalar feedback. %
Following differentiable reward fine-tuning for diffusion~\citep{clark2024directly,prabhudesai2023aligning}, we maximize $R$ through the decoder:
\begin{equation}
\mathcal{L}_{\mathrm{reward}}(\phi) =
- \,\mathbb{E}_{\hat z \sim \pi_\theta}\!\left[\,R\!\left(D_\phi(\hat z),\,y\right)\,\right].
\end{equation}
Crucially, $\hat z$ is policy-sampled and \emph{detached}: no gradient crosses the discrete boundary, so this channel never attempts the impossible task of differentiating through categorical sampling. %

\textbf{Black-box channel: Rank-GAN.}
When $R$ is non-differentiable, the channel above vanishes. %
A vanilla GAN loss~\citep{goodfellow2014generative} on policy-sampled latents would treat every rollout uniformly and discard the per-sample ranking that the policy has just been optimized over. %
We therefore introduce a reward-weighted variant inspired by reward-weighted regression~\citep{peters2007reinforcement,peng2019advantage}, which we call \textit{Rank-GAN}:
\begin{equation}
\mathcal{L}_{\mathrm{Rank\text{-}GAN}}(\phi) =
-\,\mathbb{E}_{\hat{z}\sim\pi_\theta}\!\left[\,
w(\hat{z})\cdot\mathrm{Disc}\!\left(D_\phi(\hat{z})\right)
\,\right],
\label{eq:rankgan}
\end{equation}
with weights $w(\hat z_i)\propto\exp(r_i/\tau)$ normalized so that $\sum_i w(\hat z_i)=G$. %
It preserves the expected gradient magnitude of a vanilla GAN while concentrating updates on policy-preferred samples, and the discriminator is trained adversarially against images $x_{\mathrm{gt}}$. %
Replacing Rank-GAN with a uniform GAN drops both CLIP and FID (\Cref{sec:experiments}), confirming that reward weighting is the active ingredient. %

\textbf{What the two channels share.}
Both channels move $D_\phi$ toward decoded images preferred by the reward model on the current policy distribution, but they sit at different points on a bias--variance trade-off: the differentiable channel offers low-variance pixel-space gradients when available, whereas Rank-GAN offers a reward-agnostic surrogate that requires only scalar feedback. %
We therefore retain a small weight $\lambda_d$ on $\mathcal{L}_{\mathrm{reward}}$ even when CLIP-style gradients are available; the ablation in \Cref{subsec:ablation} confirms that the combination outperforms either channel alone. %

\subsubsection{Manifold-anchored regularization ($\Omega_\phi$)}
\label{subsubsec:decoder_regularization}

Alignment alone is unsafe: trained only on stochastic policy latents under adversarial pressure, the decoder would readily abandon the deterministic ground-truth manifold on which it was originally fit---the renderer-side analogue of reward hacking. %
Two regularizers prevent this drift, each targeting a distinct failure mode. %

\textbf{Anchor 1: reconstruction on ground-truth codes.}
We retain the original tokenizer-training objective on ground-truth codes $z_{\mathrm{gt}}$~\citep{esser2021taming}:
\begin{equation}
\mathcal{L}_{\mathrm{recon}}(\phi) =
\big\|\,x_{\mathrm{gt}} - D_\phi(z_{\mathrm{gt}})\,\big\|_1
\;+\;
\mathcal{L}_{\mathrm{GAN}}\!\left(D_\phi(z_{\mathrm{gt}})\right).
\end{equation}
Mixed into every M-step, this term preserves fidelity on the deterministic ground-truth distribution against catastrophic forgetting~\citep{kirkpatrick2017overcoming} induced by training on stochastic policy samples. %

\textbf{Anchor 2: EMA-consistent stability on policy codes.}
A subtler concern is local stability. %
VQ codes lie on a discrete manifold where a single index change can produce a large pixel jump, making $D_\phi$ non-Lipschitz under stochastic sampling. %
To smooth its response in novel code regions, we distill from a slow-moving EMA teacher $D_{\phi_{\mathrm{ema}}}$~\citep{zhang2018unreasonable}, following the self-distillation paradigm~\citep{tarvainen2017mean,grill2020bootstrap}:
\begin{equation}
\mathcal{L}_{\mathrm{consist}}(\phi) =
\mathbb{E}_{\hat{z}\sim\pi_\theta}\!\left[\,
\mathcal{L}_{\mathrm{LPIPS}}\!\big(\,
D_\phi(\hat{z}),\;
\mathrm{sg}\!\left[D_{\phi_{\mathrm{ema}}}(\hat{z})\right]
\big)\,\right].
\end{equation}
The teacher provides a stable target that filters out the high-frequency noise of single-step adversarial updates on a discrete input. %
The two anchors are complementary---$\mathcal{L}_{\mathrm{recon}}$ guards against manifold forgetting on ground-truth codes, while $\mathcal{L}_{\mathrm{consist}}$ guards against over-fitting to whatever the policy happens to sample on a given step---and together they bound the decoder to a neighborhood of the pre-training manifold within which the alignment block of \Cref{subsubsec:decoder_alignment} is free to operate. %

\section{Experiments}
\label{sec:experiments}

Our experiments address a core question: under fixed reward, data, and compute, does co-evolving the VQ decoder with the AR policy yield measurable gains over the frozen-decoder convention? %
We establish a controlled setting (\Cref{subsec:setup}), conduct three demanding tests (\Cref{subsec:main_results}), verify the underlying mechanism (\Cref{subsec:mechanism}), and isolate component contributions (\Cref{subsec:ablation}). %

\subsection{Experimental Setup}
\label{subsec:setup}

\paragraph{Models and baselines.}
We evaluate RankE on two representative discrete AR T2I backbones: LlamaGen-XL~\citep{sun2024llamagen} ($775$M) and Janus-Pro-1B~\citep{deepseek2025januspro}, a unified multimodal architecture. %
We compare against three baselines of increasing post-training intensity: the pre-trained model (Base), a supervised fine-tuned variant on our curated corpus (SFT), and a standard RL baseline that updates the AR policy via GRPO~\citep{shao2024deepseekmath} under a CLIP~\citep{radford2021learning} or HPSv2~\citep{wu2023hpsv2} reward while keeping the decoder frozen (Std.\,RL). %
This last baseline is our apples-to-apples comparison: with reward, data, and total compute fixed, the only difference from RankE is whether the decoder co-evolves, so any measured gap is directly attributable to decoder adaptation alone. %

\paragraph{Datasets and evaluation.}
We evaluate along three complementary axes: fidelity and alignment through FID~\citep{heusel2017gans} and CLIP Score~\citep{radford2021learning} on MS-COCO 30K~\citep{lin2014microsoft}; human preference through HPSv2~\citep{wu2023hpsv2} on the Photo, Concept, and Anime subsets; and compositional reasoning on zero-shot GenEval~\citep{ghosh2023geneval} (Two-Object, Counting, Color binding), on which models receive no task-specific supervision, giving a direct test of generalization beyond surface-level text matching. %
The 15K training corpus is curated from BLIP3o-60k with caption compression and stratified domain sampling, and full details are given in Appendix~\ref{app:data}. %

\definecolor{oursblue}{RGB}{232, 243, 255}
\definecolor{headgray}{RGB}{245, 245, 245}
\definecolor{darkteal}{RGB}{0, 100, 80}
\definecolor{lightgraytext}{RGB}{190, 190, 190}
 
\begin{table}[t]
    \centering
    \caption{%
    Comparison of representative T2I alignment methods. %
    We focus on whether the decoder (VAE or VQ) is updated during post-training (\textbf{Dec.}). %
    Most methods keep the decoder frozen; RankE is, to our knowledge, the first discrete AR method to co-evolve the decoder. %
    FID and CLIP are reported on MS-COCO 30K where available; ``--'' indicates the metric is not reported by the original work under a comparable protocol. %
    Numbers from prior work are cited as-is and may use different reward signals or evaluation pipelines, thus not apples-to-apples; more comparison appears in \Cref{tab:results_clip}. %
    Methods besides SFT and RL baselines use different training protocols, hindering direct comparison. %
    }
    \label{tab:comparison_survey}
    \vspace{1mm}
    \resizebox{\textwidth}{!}{%
    \begin{tabular}{@{}llccllcc@{}}
    \toprule
    \textbf{Method}
      & \textbf{Venue}
      & \textbf{Size}
      & \textbf{Type}
      & \textbf{Strategy}
      & \textbf{Dec.}
      & \textbf{FID}$\downarrow$
      & \textbf{CLIP}$\uparrow$ \\
    \midrule
 
    \multicolumn{8}{l}{\cellcolor{headgray}
      \textit{\textbf{Diffusion \& Flow Matching}}} \\[2pt]
    SDXL~\cite{Podell2023SDXLIL}
      & arXiv'23 & 2.6B & Diff.
      & Pretrain & Frozen & -- & 31.0 \\
    DPOK~\cite{fan2023dpok}
      & NeurIPS'23 & 0.9B & Diff.
      & RL + Direct BP & Frozen & -- & -- \\
    Diffusion-DPO~\cite{wallace2023diffusion}
      & CVPR'24 & 2.6B & Diff.
      & DPO & Frozen & 10.5 & 31.5 \\
    REPA-E~\cite{leng2025repa}
      & CVPR'25 & LoRA & Diff.
      & E2E Reparam. & Unfrozen & -- & -- \\
    \addlinespace[3pt]

    Flow-GRPO~\cite{Liu2025FlowGRPO}
      & NeurIPS'25 & 2.0B & Flow
      & Online GRPO & Frozen & 10.5 & -- \\
 
    \multicolumn{8}{l}{\cellcolor{headgray}
      \textit{\textbf{Autoregressive (AR) \& Unified}}} \\[2pt]
    LlamaGen~\cite{sun2024llamagen}
      & arXiv'24 & 775M & AR
      & Pretrain & Unfrozen & 15.24 & 31.54 \\

    Transfusion~\cite{Zhou2024Transfusion}
      & ICLR'25 & 7.0B & Unified
      & Mixed-Modal & Frozen & 16.9 & 25.5 \\
    Emu3~\cite{wang2024emu3}
      & arXiv'24 & 7.0B & AR
      & Unified AR & Frozen & 11.6 & 28.6 \\
    Janus-Pro~\cite{deepseek2025januspro}
      & arXiv'25 & 7.0B & Uni.
      & Decoupled Enc. & Frozen & 6.85 & 32.0 \\
    \addlinespace[3pt]
      
    GCPO~\cite{zhang2026group}
      & ICLR'26 & 775M & AR
      & Critical-Token Opt. & Frozen & -- & -- \\

    VA-$\pi$~\cite{liao2025va}
      & CVPR'26 & 775M & AR
      & Variational Align. & Frozen & 9.23 & 29.1 \\
 
    \arrayrulecolor{black!25}\midrule\arrayrulecolor{black}
 
    LlamaGen-SFT
      & -- & 775M & AR
      & SFT & Frozen & 16.58 & 31.86 \\
    LlamaGen-RL
      & -- & 775M & AR
      & GRPO & Frozen & 17.76 & 32.45 \\
 
    \rowcolor{oursblue}
    \textbf{RankE (Ours)}
      & -- & \textbf{775M} & \textbf{AR}
      & \textbf{Co-Evolved} & \textbf{Unfrozen}
      & \textbf{15.21} & \textbf{33.76} \\
 
    \bottomrule
    \end{tabular}%
    }
    
    \vspace{0.5em}
    \footnotesize

\end{table}

\subsection{Does Decoder Co-Evolution Help? Three Controlled Tests}
\label{subsec:main_results}

We test our central question with three lenses, moving from the broadest to the most stringent. %

\paragraph{Test 1: positioning across the post-training landscape.}
Table~\ref{tab:comparison_survey} situates RankE within the broader post-training picture, spanning diffusion models~\citep{rombach2022high,peebles2023dit}, flow matching~\citep{lipman2022flow}, unified architectures~\citep{tian2024visual}, and AR generators~\citep{yu2022parti}. %
A clear pattern emerges: on the diffusion side, recent methods such as DRaFT~\citep{clark2024directly} and REPA-E~\citep{leng2025repa} have begun to unlock the renderer for joint optimization; on the AR side, alignment-focused methods including GCPO~\citep{zhang2026group} and VA-$\pi$~\citep{liao2025va} still freeze the decoder. %
RankE is, to the best of our knowledge, the first AR method to co-evolve the decoder with the policy. %
Among methods that explicitly target the CLIP/FID alignment trade-off on MS-COCO 30K, RankE attains a favorable operating point of FID $15.21$ and CLIP $33.76$ at $775$M parameters; comparisons that report different reward signals or evaluation protocols are not directly commensurable. %

\definecolor{gaingreen}{RGB}{0,150,0}
\newcommand{\gain}[1]{\textcolor{gaingreen}{\scriptsize(#1)}}
\newcommand{\gaingood}[1]{\textcolor{gaingreen}{\scriptsize($\uparrow$#1)}}
\newcommand{\gainbetter}[1]{\textcolor{gaingreen}{\scriptsize($\downarrow$#1)}}

\begin{table}[t!]
    \centering
    \caption{%
            Quantitative results under CLIP-based optimization.
            The standard RL baseline improves CLIP score but degrades
            image fidelity.
            RankE co-evolves the decoder with the policy, achieving
            higher CLIP score and lower FID,
            demonstrating that decoder adaptation converts reward
            improvement into pixel-space gains without sacrificing fidelity.
            \textcolor{gaingreen}{Green} numbers denote gains over Std.\,RL.
        }
    \label{tab:results_clip}
    \renewcommand{\arraystretch}{1.2}
    \resizebox{1\textwidth}{!}{%
        \begin{tabular}{@{} l l c cc cccc @{}}
            \toprule
            \multirow{2}{*}{\textbf{Backbone}} &
            \multirow{2}{*}{\textbf{Method (Reward)}} &
            \multirow{2}{*}{\textbf{Decoder}} &
            \multicolumn{2}{c}{\textbf{MS-COCO}} &
            \multicolumn{4}{c}{\textbf{GenEval (Zero-shot)}} \\
            \cmidrule(lr){4-5} \cmidrule(l){6-9}
            & & &
            \textbf{CLIP}\,$\uparrow$ & \textbf{FID}\,$\downarrow$ &
            2-Obj & Count & Color & \textbf{Avg.}\,$\uparrow$ \\
            \midrule
            \multirow{4}{*}{%
                \begin{tabular}[c]{@{}l@{}}
                    \textbf{LlamaGen-XL}\\(775M)
                \end{tabular}}
            & Base
                & Frozen
                & 31.54 & 15.24
                & 0.331 & 0.175 & 0.559 & 0.309 \\
            \hdashline\noalign{\vskip 0.5ex}
            & SFT
                & Frozen
                & 31.86 & 16.58
                & 0.288 & 0.228 & 0.723 & 0.374 \\
            & Std.\ RL (\textit{CLIP})
                & Frozen
                & 32.45 & 17.76
                & 0.361 & 0.272 & 0.779 & 0.417 \\
            \rowcolor{oursblue}\cellcolor{white}
            & \textbf{RankE (\textit{CLIP})}
                & \textbf{Co-Evol}
                & \textbf{33.76} \gaingood{1.31} & \textbf{15.21} \gainbetter{2.55}
                & 0.379  & 0.356 & 0.763 & \textbf{0.425} \gaingood{.008} \\
            \midrule
            \multirow{4}{*}{%
                \begin{tabular}[c]{@{}l@{}}
                    \textbf{Janus-Pro}\\(1B)
                \end{tabular}}
            & Base
                & Frozen
                & 33.20 & 18.95
                & 0.810 & 0.553 & 0.894 & 0.740 \\
            \hdashline\noalign{\vskip 0.5ex}
            & SFT
                & Frozen
                & 33.31 & 26.73
                & 0.826 & 0.572 & 0.889 & 0.739 \\
            & Std.\ RL (\textit{CLIP})
                & Frozen
                & 33.60 & 25.59
                & 0.838 & 0.478 & 0.891 & 0.746 \\
            \rowcolor{oursblue}\cellcolor{white}
            & \textbf{RankE (\textit{CLIP})}
                & \textbf{Co-Evol}
                & \textbf{33.86} \gaingood{0.26} & \textbf{25.19} \gainbetter{0.40}
                & 0.836 & 0.487 & 0.892 & \textbf{0.750} \gaingood{.004} \\
            \bottomrule
        \end{tabular}%
    }
\end{table}

\begin{table}[t!]
    \centering
\caption{%
        Quantitative results under HPSv2-based optimization.
        RankE co-evolves the decoder with the autoregressive policy,
        improving preference alignment in pixel space while preserving
        generation fidelity. RankE also maintains strong zero-shot
        compositional performance on GenEval, indicating that the
        alignment gains do not compromise generalization across
        diverse visual reasoning tasks.
    }
    \label{tab:results_hpsv2}
    \renewcommand{\arraystretch}{1.2}
    \resizebox{\textwidth}{!}{%
        \begin{tabular}{@{} l l c cccc cccc @{}}
            \toprule
            \multirow{2}{*}{\textbf{Backbone}} &
            \multirow{2}{*}{\textbf{Method (Reward)}} &
            \multirow{2}{*}{\textbf{Decoder}} &
            \multicolumn{4}{c}{\textbf{HPSv2}} &
            \multicolumn{4}{c}{\textbf{GenEval (Zero-shot)}} \\
            \cmidrule(lr){4-7} \cmidrule(l){8-11}
            & & &
            Photo & Concept & Anime & \textbf{Avg.}\,$\uparrow$ &
            2-Obj & Count & Color & \textbf{Avg.}\,$\uparrow$ \\
            \midrule
            \multirow{4}{*}{%
                \begin{tabular}[c]{@{}l@{}}
                    \textbf{LlamaGen-XL}\\(775M)
                \end{tabular}}
            & Base
                & Frozen
                & 0.2364 & 0.2107 & 0.2183 & 0.2196
                & 0.331  & 0.175  & 0.559  & 0.309 \\
            \hdashline\noalign{\vskip 0.5ex}
            & SFT
                & Frozen
                & 0.2281 & 0.2202 & 0.2222 & 0.2221
                & 0.288  & 0.228  & 0.723  & 0.374 \\
            & Std.\ RL (\textit{HPSv2})
                & Frozen
                & 0.2466 & 0.2435 & 0.2436 & 0.2451
                & 0.397  & 0.259  & 0.798  & 0.418 \\
            \rowcolor{oursblue}\cellcolor{white}
            & \textbf{RankE (\textit{HPSv2})}
                & \textbf{Co-Evol}
                & \textbf{0.2492} & \textbf{0.2479} & \textbf{0.2453} & \textbf{0.2531}
                & 0.386 & \textbf{0.291} & 0.771 & \textbf{0.423} \\
            \bottomrule
        \end{tabular}%
    }
\end{table}

\begin{figure}[h!]
    \centering
    \includegraphics[width=\textwidth]{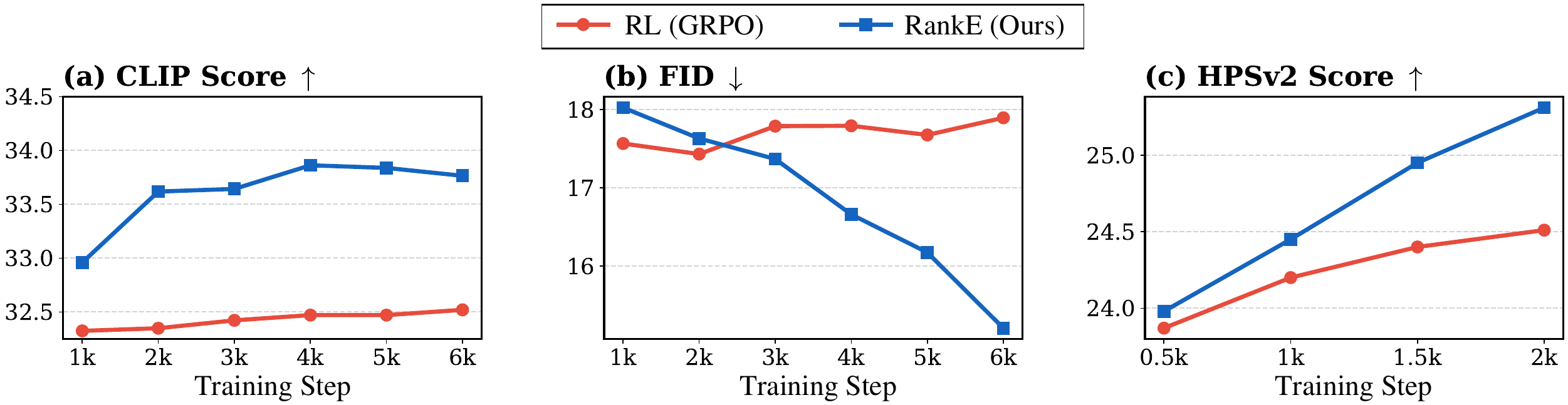}
    \caption{\textbf{Evolution of generation metrics over training steps.} Comparison of RankE against a Standard RL baseline with a frozen decoder. While Standard RL achieves marginal improvements in alignment, it suffers from stagnant or degrading visual fidelity (b) as the frozen decoder cannot adapt to policy-induced latent drift. In contrast, the co-evolution mechanism of RankE effectively translates reward optimization into pixel-space gains, simultaneously improving semantic alignment (a), image fidelity (b), and human preference scores (c).}
    \label{fig:related_teaser}
    \vspace{-4mm}
\end{figure}

\paragraph{Test 2: apples-to-apples controlled comparison.}
Under a CLIP reward (\Cref{tab:results_clip}), standard RL improves CLIP but degrades FID (LlamaGen-XL: $16.58\!\to\!17.76$), as the frozen decoder cannot track drifting latents. %
RankE reverses this, reaching FID $15.21$ and CLIP $33.76$. %
On Janus-Pro-1B, RankE yields the best CLIP and zero-shot GenEval; all post-training variants here regress FID vs.\ Base, likely due to a corpus-matching limitation (\Cref{sec:conclusion}). %
Trajectory-wise (\Cref{fig:related_teaser}), RankE monotonically improves both metrics, whereas standard RL degrades FID. %
This generalizes to non-differentiable rewards (\Cref{tab:results_hpsv2}): RankE boosts HPSv2 from $0.2451$ to $0.2531$ while maintaining GenEval performance, showing that gains stem from decoder adaptation rather than from specific reward types. %

\begin{figure}[t]
    \centering
    \includegraphics[width=\textwidth]{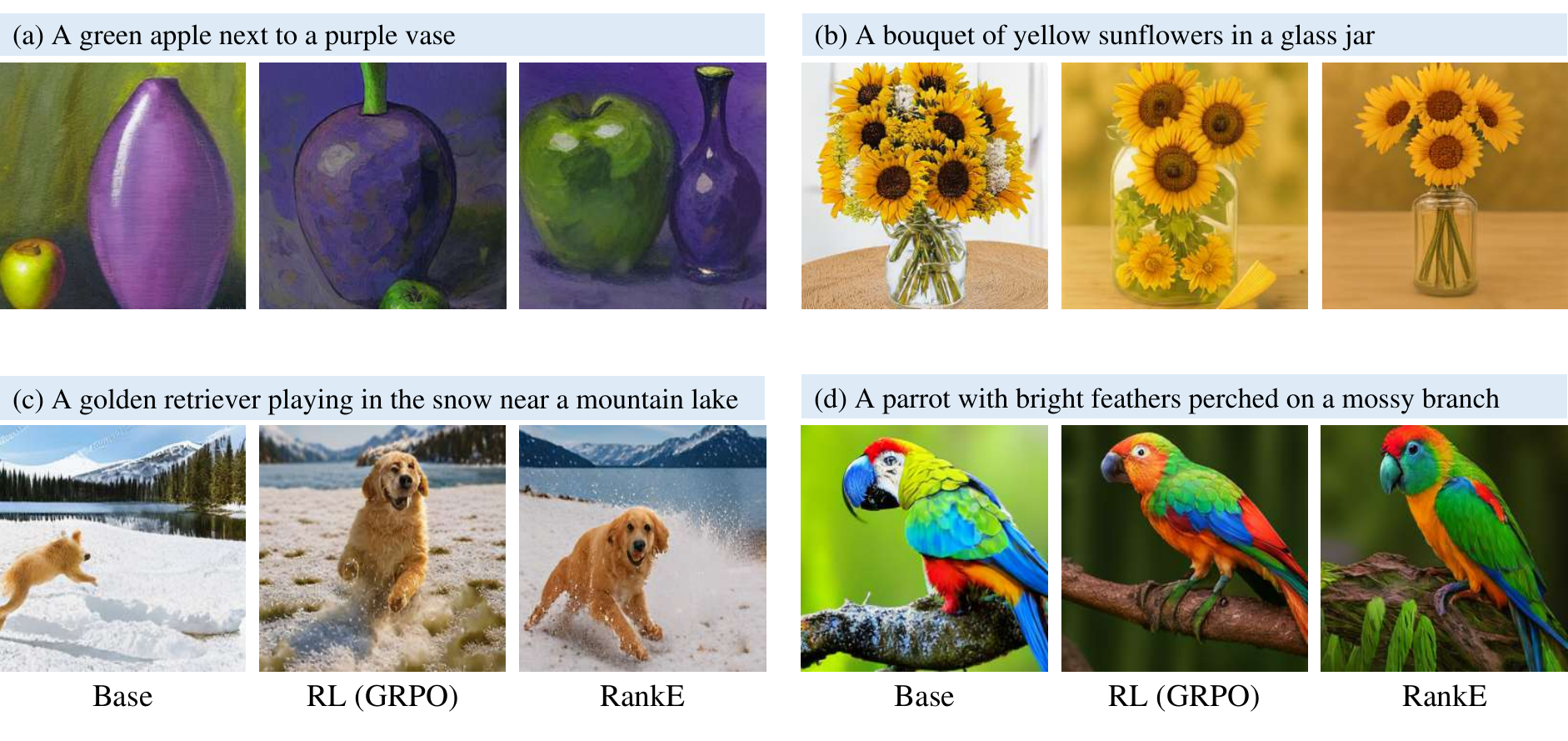}
    \caption{Visualization of T2I generation. RankE yields precise attributes and details according to the text prompt when compared to the baseline, while reducing the artifacts with high image quality in comparison to the existing RL method (GRPO).}
    \label{fig:exp_result_visualize}
\end{figure}

\paragraph{Test 3: qualitative verification.}
\Cref{fig:exp_result_visualize} compares generations under matched prompts. %
The base model frequently misses prompt attributes (color, count, spatial relations); standard RL improves adherence at the cost of visible artifacts, a direct consequence of the frozen decoder processing latents drawn from a distribution it was never trained on; and RankE produces images with both faithful attributes and high perceptual quality, with none of the artifact bands of the standard RL baseline. %
Decoder adaptation, in other words, is what turns latent-space alignment into pixel-space fidelity. %

\begin{figure}[t]
    \centering
    \includegraphics[width=0.95\linewidth]{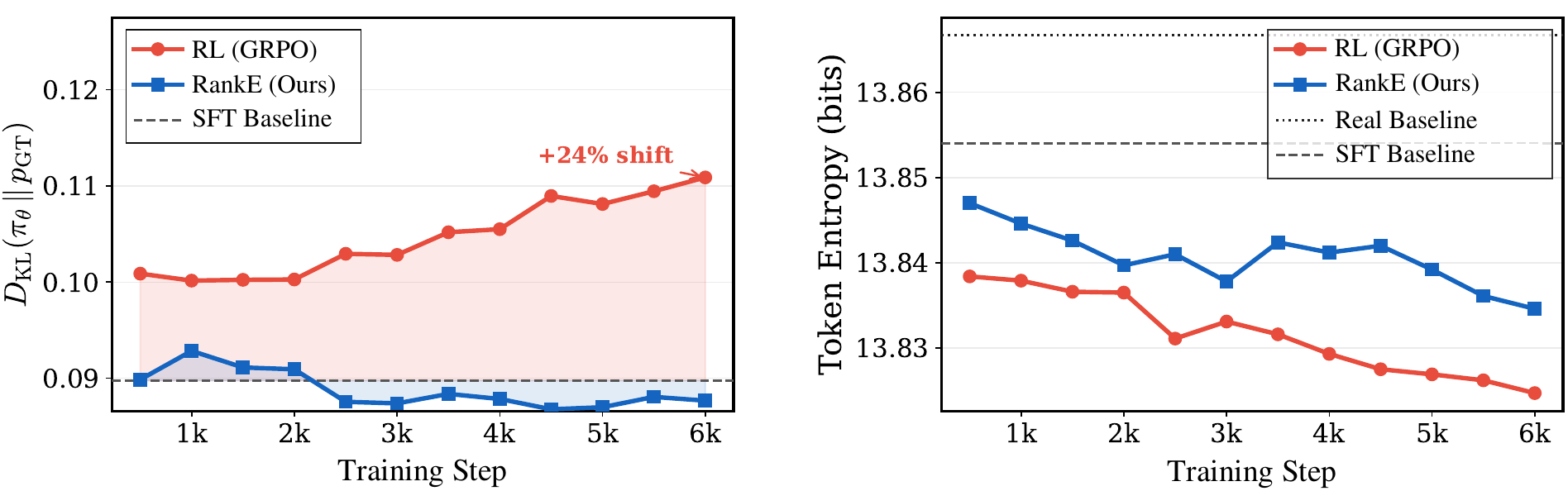}
    \caption{Latent Covariate Shift and token entropy during training.
    \textbf{Left:} distributional shift measured as
    $D_{\mathrm{KL}}(\pi_\theta \,\|\, p_{\mathrm{real}})$,
    evaluated every 500 steps. Standard RL with a frozen decoder
    shows steadily increasing divergence ($+24\%$ over training),
    whereas RankE keeps the divergence near the SFT
    initialization.
    \textbf{Right:} standard RL reduces token entropy as the
    policy concentrates on fewer codebook entries; RankE maintains
    entropy closer to the real-image level ($13.87$ bits).}
    \label{fig:analyse_combine_shift_entropy}
    \vspace{-4mm}
\end{figure}

\subsection{Why Does It Work? Diagnosing the Mechanism}
\label{subsec:mechanism}

Having established that decoder co-evolution works, we verify whether the mechanism matches our hypothesis in \Cref{sec:intro}. %
If RankE truly absorbs Latent Covariate Shift rather than relying on unrelated regularization, two predictions follow. %
First, the KL divergence $D_{\mathrm{KL}}(\pi_\theta\|p_{\mathrm{real}})$ between policy tokens and the training distribution of the tokenizer should stay bounded under RankE while diverging under frozen-decoder RL. %
Second, codebook entropy should remain close to the real-image level rather than collapsing onto a few favored indices. %

\Cref{fig:analyse_combine_shift_entropy} confirms both on LlamaGen-XL. %
The left panel shows that the KL divergence under standard RL rises steadily by $+24\%$, exactly the drift predicted in \Cref{sec:intro}, while RankE stays at or slightly below the SFT initialization across the run. %
The right panel tells the symmetric story in entropy: standard RL concentrates the policy on fewer codebook entries, which intensifies the mismatch with the decoder pre-training distribution, while RankE preserves entropy near the real-image level ($13.87$ bits). %
The two diagnostics provide direct evidence that co-evolution absorbs Latent Covariate Shift during training, rather than papering over its downstream consequences. %

\begin{table}[h]
\vspace{-1em}
\centering
\footnotesize
\caption{%
    Effect of training mode (CLIP reward, MS-COCO 30K). %
    Full joint training outperforms all partial configurations, confirming the synergistic benefit of co-evolving both modules. %
}
\label{tab:abl_train_mode}
\setlength{\tabcolsep}{3pt}
\renewcommand{\arraystretch}{1.1}
\begin{tabular}{@{} l cc ccc @{}}
\toprule
\textbf{Setting}
  & \textbf{Policy} & \textbf{Decoder}
  & \textbf{CLIP}\,$\uparrow$
  & \textbf{FID}\,$\downarrow$
  & \textbf{GenEval}\,$\uparrow$ \\
\midrule
SFT (No Post-Train) & \xmark & \xmark & 31.86 & 16.58 & 0.374 \\
Policy-Only (GRPO)  & \cmark & \xmark & 32.45 & 17.76 & 0.417 \\
Decoder-Only         & \xmark & \cmark & 33.41 & 18.68 & 0.403 \\
\rowcolor{oursblue}
\textbf{Full RankE}  & \cmark & \cmark
  & \textbf{33.76} & \textbf{15.21} & \textbf{0.425} \\
\bottomrule
\end{tabular}
\vspace{-2mm}
\end{table}

\subsection{What Drives the Gains? Component-Level Ablation}
\label{subsec:ablation}

Having established that decoder co-evolution works and operates through the predicted mechanism, we isolate which design choices are responsible. %
We ablate two design axes on LlamaGen-XL under a CLIP reward on MS-COCO 30K: the high-level training mode and the composition of the decoder loss. %
Further sensitivity analyses appear in Appendix~\ref{app:extended_ablation}. %

\paragraph{Training mode: is co-evolution synergistic, or just the union of two effects?}
\Cref{tab:abl_train_mode} compares four configurations. %
Policy-only GRPO (Row~2) improves CLIP but degrades FID, which is the Latent Covariate Shift diagnosis again. %
Decoder-only adaptation (Row~3) is more interesting: it raises CLIP to $33.41$ without any policy-level signal, which exposes a decoder-side gap that the existing literature has overlooked; FID nonetheless degrades to $18.68$ without coordinated policy guidance. %
Only full RankE (Row~4) is best on both fidelity (FID) and overall composition (GenEval), and matches the best CLIP. %
On FID in particular, full RankE ($15.21$) improves substantially over either policy-only ($17.76$) or decoder-only ($18.68$), indicating that the two updates jointly absorb Latent Covariate Shift in a way neither alone can. %

\begin{table}[h]
\vspace{-2mm}
\centering
\footnotesize
\caption{%
    Contribution of each decoder loss component under joint training (CLIP reward, MS-COCO 30K). %
    Each term is removed individually from the full configuration to measure its marginal effect. %
}
\label{tab:abl_dec_loss}
\setlength{\tabcolsep}{3.5pt}
\renewcommand{\arraystretch}{1.1}
\begin{tabular}{@{} l cccc ccc @{}}
\toprule
\textbf{Setting}
  & $\mathcal{L}_\text{r}$ & $\mathcal{L}_\text{g}$
  & $\mathcal{L}_\text{c}$ & $\mathcal{L}_\text{d}$
  & \textbf{CLIP}\,$\uparrow$
  & \textbf{FID}\,$\downarrow$
  & \textbf{GenEval}\,$\uparrow$ \\
\midrule
w/o Reconstruction
  & \xmark & \cmark & \cmark & \cmark & 33.26 & 17.69 & 0.401 \\
w/o GAN
  & \cmark & \xmark & \cmark & \cmark & 32.12 & 18.59 & 0.397 \\
w/o Consistency
  & \cmark & \cmark & \xmark & \cmark & 34.17 & 19.03 & 0.408 \\
w/o Reward BP & \cmark & \cmark & \cmark & \xmark & 32.53 & 20.16 & 0.412 \\
\rowcolor{oursblue}
\textbf{Full RankE}
  & \cmark & \cmark & \cmark & \cmark
  & \textbf{33.76} & \textbf{15.21} & \textbf{0.425} \\
\bottomrule
\end{tabular}
\end{table}

\paragraph{Decoder loss ablation: which term inside $\mathcal{L}_D$ matters?}
\Cref{tab:abl_dec_loss} removes each term individually, which separates the alignment block from the regularization block of Eq.~\ref{eq:decoder}. %
The pattern aligns exactly with the roles assigned in \Cref{subsec:decoder_design}. %
Removing $\mathcal{L}_{\mathrm{Rank\text{-}GAN}}$ drops both CLIP and FID, which confirms it as the primary channel for non-differentiable reward signals; removing $\mathcal{L}_{\mathrm{reward}}$ shows that even small differentiable gradients contribute measurable additional gains beyond Rank-GAN alone, exactly the bias--variance complementarity argued in \Cref{subsubsec:decoder_alignment}. %
On the regularization side, removing $\mathcal{L}_{\mathrm{recon}}$ eliminates the ground-truth anchor and FID degrades to $17.69$ (manifold forgetting), while removing $\mathcal{L}_{\mathrm{consist}}$ exhibits the asymmetric failure mode predicted in \Cref{subsubsec:decoder_regularization}: CLIP rises slightly to $34.17$ but FID degrades to $19.03$, because the decoder is now free to over-fit to whatever instantaneous policy latents it sees. %
Each of the four terms maps onto a distinct failure mode, and removing any one of them re-opens precisely the hole it was designed to plug. %

\section{Conclusion}
\label{sec:conclusion}

\paragraph{Conclusion.}
We identify \emph{Latent Covariate Shift} as a fundamental bottleneck in the post-training of discrete AR T2I models: the VQ decoder is trained on deterministically quantized ground-truth codes, whereas inference relies on tokens sampled from a stochastic, reward-driven policy. %
To absorb this shift rather than accumulate it, we propose \textbf{RankE}, the first end-to-end post-training framework that co-evolves the AR policy and the VQ decoder, instantiated as a Generalized EM procedure over the latent--pixel chain. %
Across two backbones (LlamaGen-XL, Janus-Pro), three evaluation axes (FID, CLIP/HPSv2, GenEval), and two reward functions, RankE consistently overcomes the Pareto trade-off of frozen-decoder baselines by improving both fidelity and alignment simultaneously. %

\paragraph{Limitations and future work.}
Three limitations frame our results and indicate natural directions for future work. %
\textit{(i) Memory footprint and scheduling.}
Although the temporal overhead is marginal, holding the discriminator and EMA decoder increases peak VRAM compared with single-stage post-training (see Appendix~\ref{app:impl}); adaptive scheduling triggered by reward plateaus or less frequent decoder updates are natural ways to further optimize resource efficiency. %
\textit{(ii) Sensitivity to SFT corpus alignment.}
The gains of RankE depend on the alignment between the SFT corpus and the backbone pre-training distribution. %
On Janus-Pro, pre-trained on an inaccessible proprietary corpus, SFT alone regresses FID from $18.95$ to $26.73$, capping the headroom available for post-training; RankE still improves over Std.\,RL on every metric, without surpassing the Base FID on this backbone. %
We characterize this as a corpus-matching limitation rather than a methodological flaw of co-evolution. %
\textit{(iii) Frozen encoder.}
We freeze the VQ encoder so that ground-truth tokens remain a stable anchor for $\mathcal{L}_{\mathrm{recon}}$; jointly training the encoder, extending co-evolution to pre-training, integrating online human feedback~\citep{ouyang2022training}, and exploring multi-objective reward composition are natural next steps. %

{
\small
\bibliographystyle{plainnat}
\bibliography{main}

\begin{thebibliography}{73}
\providecommand{\natexlab}[1]{#1}
\providecommand{\url}[1]{\texttt{#1}}
\expandafter\ifx\csname urlstyle\endcsname\relax
  \providecommand{\doi}[1]{doi: #1}\else
  \providecommand{\doi}{doi: \begingroup \urlstyle{rm}\Url}\fi

\bibitem[Bengio et~al.(2015)Bengio, Vinyals, Jaitly, and Shazeer]{bengio2015scheduled}
Samy Bengio, Oriol Vinyals, Navdeep Jaitly, and Noam Shazeer.
\newblock Scheduled sampling for sequence prediction with recurrent neural networks.
\newblock In \emph{NeurIPS}, 2015.

\bibitem[Bengio et~al.(2013)Bengio, L{\'e}onard, and Courville]{bengio2013estimating}
Yoshua Bengio, Nicholas L{\'e}onard, and Aaron Courville.
\newblock Estimating or propagating gradients through stochastic neurons for conditional computation.
\newblock \emph{arXiv preprint arXiv:1308.3432}, 2013.

\bibitem[Betker et~al.(2023)Betker, Goh, Jing, Brooks, Wang, Li, Ouyang, Zhuang, Lee, Guo, et~al.]{betker2023improving}
James Betker, Gabriel Goh, Li~Jing, Tim Brooks, Jianfeng Wang, Linjie Li, Long Ouyang, Juntang Zhuang, Joyce Lee, Yufei Guo, et~al.
\newblock Improving image generation with better captions.
\newblock \emph{Computer Science}, 2023.

\bibitem[Black et~al.(2024)Black, Janner, Du, Kostrikov, and Levine]{black2024training}
Kevin Black, Michael Janner, Yilun Du, Ilya Kostrikov, and Sergey Levine.
\newblock Training diffusion models with reinforcement learning.
\newblock In \emph{ICLR}, 2024.

\bibitem[{Chameleon Team}(2024)]{team2024chameleon}
{Chameleon Team}.
\newblock Chameleon: Mixed-modal early-fusion foundation models.
\newblock \emph{arXiv preprint arXiv:2405.09818}, 2024.

\bibitem[Chang et~al.(2022)Chang, Zhang, Jiang, Liu, and Freeman]{chang2022maskgit}
Huiwen Chang, Han Zhang, Lu~Jiang, Ce~Liu, and William~T Freeman.
\newblock {MaskGIT}: Masked generative image transformer.
\newblock In \emph{CVPR}, 2022.

\bibitem[Chang et~al.(2023)Chang, Zhang, Barber, Maschinot, Lezama, Jiang, Yang, Murphy, Freeman, Rubinstein, et~al.]{chang2023muse}
Huiwen Chang, Han Zhang, Jarred Barber, AJ~Maschinot, Jose Lezama, Lu~Jiang, Ming-Hsuan Yang, Kevin Murphy, William~T Freeman, Michael Rubinstein, et~al.
\newblock Muse: Text-to-image generation via masked generative transformers.
\newblock In \emph{ICML}, 2023.

\bibitem[Chen et~al.(2025{\natexlab{a}})Chen, Wang, Li, Sun, Chen, Liu, Wang, Raj, Liu, and Barsoum]{Chen2024SoftVQVAEE1}
Hao Chen, Ze~Wang, Xiang Li, Ximeng Sun, Fangyi Chen, Jiang Liu, Jindong Wang, Bhiksha Raj, Zicheng Liu, and Emad Barsoum.
\newblock Softvq-vae: Efficient 1-dimensional continuous tokenizer.
\newblock In \emph{CVPR}, 2025{\natexlab{a}}.

\bibitem[Chen et~al.(2025{\natexlab{b}})Chen, Xu, Pan, Hu, Qin, Goldstein, Huang, Zhou, Xie, Savarese, et~al.]{chen2025blip3}
Jiuhai Chen, Zhiyang Xu, Xichen Pan, Yushi Hu, Can Qin, Tom Goldstein, Lifu Huang, Tianyi Zhou, Saining Xie, Silvio Savarese, et~al.
\newblock Blip3-o: A family of fully open unified multimodal models-architecture, training and dataset.
\newblock \emph{arXiv preprint arXiv:2505.09568}, 2025{\natexlab{b}}.

\bibitem[Chen et~al.(2025{\natexlab{c}})Chen, Wu, Liu, Pan, Liu, Xie, Yu, and Ruan]{deepseek2025januspro}
Xiaokang Chen, Zhiyu Wu, Xingchao Liu, Zizheng Pan, Wen Liu, Zhenda Xie, Xingkai Yu, and Chong Ruan.
\newblock Janus-pro: Unified multimodal understanding and generation with data and model scaling.
\newblock \emph{arXiv preprint arXiv:2501.17811}, 2025{\natexlab{c}}.

\bibitem[Clark et~al.(2024)Clark, Vicol, Swersky, and Fleet]{clark2024directly}
Kevin Clark, Paul Vicol, Kevin Swersky, and David~J Fleet.
\newblock Directly fine-tuning diffusion models on differentiable rewards.
\newblock In \emph{ICLR}, 2024.

\bibitem[Coste et~al.(2024)Coste, Anwar, Kirk, and Krueger]{coste2023reward}
Thomas Coste, Usman Anwar, Robert Kirk, and David Krueger.
\newblock Reward model ensembles help mitigate overoptimization.
\newblock In \emph{ICLR}, 2024.

\bibitem[Dempster et~al.(1977)Dempster, Laird, and Rubin]{dempster1977maximum}
Arthur~P Dempster, Nan~M Laird, and Donald~B Rubin.
\newblock Maximum likelihood from incomplete data via the {EM} algorithm.
\newblock \emph{Journal of the Royal Statistical Society: Series B}, 1977.

\bibitem[Ding et~al.(2021)Ding, Yang, Hong, Zheng, Zhou, Yin, Lin, Zou, Shao, Yang, and Tang]{ding2021cogview}
Ming Ding, Zhuoyi Yang, Wenyi Hong, Wendi Zheng, Chang Zhou, Da~Yin, Junyang Lin, Xu~Zou, Zhou Shao, Hongxia Yang, and Jie Tang.
\newblock {CogView}: Mastering text-to-image generation via transformers.
\newblock In \emph{NeurIPS}, 2021.

\bibitem[Esser et~al.(2021)Esser, Rombach, and Ommer]{esser2021taming}
Patrick Esser, Robin Rombach, and Bjorn Ommer.
\newblock Taming transformers for high-resolution image synthesis.
\newblock In \emph{CVPR}, 2021.

\bibitem[Fan et~al.(2024)Fan, Watkins, Du, Liu, Ryu, Boutilier, Abbeel, Ghavamzadeh, Lee, and Lee]{fan2023dpok}
Ying Fan, Olivia Watkins, Yuqing Du, Hao Liu, Moonkyung Ryu, Craig Boutilier, Pieter Abbeel, Mohammad Ghavamzadeh, Kangwook Lee, and Kimin Lee.
\newblock Dpok: Reinforcement learning for fine-tuning text-to-image diffusion models.
\newblock In \emph{NeurIPS}, 2024.

\bibitem[Gao et~al.(2023)Gao, Schulman, and Hilton]{gao2023scaling}
Leo Gao, John Schulman, and Jacob Hilton.
\newblock Scaling laws for reward model overoptimization.
\newblock In \emph{ICML}, 2023.

\bibitem[Ghosh et~al.(2023)Ghosh, Hajishirzi, and Schwing]{ghosh2023geneval}
Dhruba Ghosh, Hannaneh Hajishirzi, and Alexander Schwing.
\newblock {GenEval}: An object-focused framework for evaluating text-to-image alignment.
\newblock \emph{arXiv preprint arXiv:2310.11513}, 2023.

\bibitem[Goodfellow et~al.(2014)Goodfellow, Pouget-Abadie, Mirza, Xu, Warde-Farley, Ozair, Courville, and Bengio]{goodfellow2014generative}
Ian Goodfellow, Jean Pouget-Abadie, Mehdi Mirza, Bing Xu, David Warde-Farley, Sherjil Ozair, Aaron Courville, and Yoshua Bengio.
\newblock Generative adversarial nets.
\newblock In \emph{NeurIPS}, 2014.

\bibitem[Grill et~al.(2020)Grill, Strub, Altch{\'e}, Tallec, Richemond, Buchatskaya, Doersch, Avila~Pires, Guo, Gheshlaghi~Azar, et~al.]{grill2020bootstrap}
Jean-Bastien Grill, Florian Strub, Florent Altch{\'e}, Corentin Tallec, Pierre Richemond, Elena Buchatskaya, Carl Doersch, Bernardo Avila~Pires, Zhaohan Guo, Mohammad Gheshlaghi~Azar, et~al.
\newblock Bootstrap your own latent: A new approach to self-supervised learning.
\newblock In \emph{NeurIPS}, 2020.

\bibitem[Heusel et~al.(2017)Heusel, Ramsauer, Unterthiner, Nessler, and Hochreiter]{heusel2017gans}
Martin Heusel, Hubert Ramsauer, Thomas Unterthiner, Bernhard Nessler, and Sepp Hochreiter.
\newblock Gans trained by a two time-scale update rule converge to a local nash equilibrium.
\newblock In \emph{NeurIPS}, 2017.

\bibitem[Huh et~al.(2023)Huh, Cheung, Agrawal, and Isola]{huh2023straightening}
Minyoung Huh, Brian Cheung, Pulkit Agrawal, and Phillip Isola.
\newblock Straightening out the straight-through estimator: Overcoming optimization challenges in vector quantized networks.
\newblock \emph{ICML}, 2023.

\bibitem[Isola et~al.(2017)Isola, Zhu, Zhou, and Efros]{isola2017image}
Phillip Isola, Jun-Yan Zhu, Tinghui Zhou, and Alexei~A Efros.
\newblock Image-to-image translation with conditional adversarial networks.
\newblock In \emph{CVPR}, 2017.

\bibitem[Jang et~al.(2017)Jang, Gu, and Poole]{jang2016categorical}
Eric Jang, Shixiang Gu, and Ben Poole.
\newblock Categorical reparameterization with {G}umbel-softmax.
\newblock In \emph{ICLR}, 2017.

\bibitem[Jiang et~al.(2025)Jiang, Guo, Zhang, Zong, Li, Zhuo, Yan, Heng, and Li]{jiang2025t2i}
Dongzhi Jiang, Ziyu Guo, Renrui Zhang, Zhuofan Zong, Hao Li, Le~Zhuo, Shilin Yan, Pheng-Ann Heng, and Hongsheng Li.
\newblock T2i-r1: Reinforcing image generation with collaborative semantic-level and token-level cot.
\newblock \emph{arXiv preprint arXiv:2505.00703}, 2025.

\bibitem[Kaiser et~al.(2018)Kaiser, Roy, Vaswani, Parmar, Bengio, Unkber, and Shazeer]{kaiser2018fast}
{\L}ukasz Kaiser, Aurko Roy, Ashish Vaswani, Niki Parmar, Samy Bengio, Jakob Unkber, and Noam Shazeer.
\newblock Fast decoding in sequence models using discrete latent variables.
\newblock In \emph{ICML}, 2018.

\bibitem[Kaplan et~al.(2020)Kaplan, McCandlish, Henighan, Brown, Chess, Child, Gray, Radford, Wu, and Amodei]{kaplan2020scaling}
Jared Kaplan, Sam McCandlish, Tom Henighan, Tom~B Brown, Benjamin Chess, Rewon Child, Scott Gray, Alec Radford, Jeffrey Wu, and Dario Amodei.
\newblock Scaling laws for neural language models.
\newblock \emph{arXiv preprint arXiv:2001.08361}, 2020.

\bibitem[Kirkpatrick et~al.(2017)Kirkpatrick, Pascanu, Rabinowitz, Veness, Desjardins, Rusu, Milan, Quan, Ramalho, Grabska-Barwinska, et~al.]{kirkpatrick2017overcoming}
James Kirkpatrick, Razvan Pascanu, Neil Rabinowitz, Joel Veness, Guillaume Desjardins, Andrei~A Rusu, Kieran Milan, John Quan, Tiago Ramalho, Agnieszka Grabska-Barwinska, et~al.
\newblock Overcoming catastrophic forgetting in neural networks.
\newblock \emph{Proceedings of the national academy of sciences}, 2017.

\bibitem[Korbak et~al.(2022)Korbak, Elsahar, Kruszewski, and Dymetman]{korbak2022rl}
Tomasz Korbak, Hady Elsahar, Germ{\'a}n Kruszewski, and Marc Dymetman.
\newblock Rl with kl penalties is better viewed as bayesian inference.
\newblock In \emph{EMNLP}, 2022.

\bibitem[Leng et~al.(2025)Leng, Singh, Hou, Xing, Xie, and Zheng]{leng2025repa}
Xingjian Leng, Jaskirat Singh, Yunzhong Hou, Zhenchang Xing, Saining Xie, and Liang Zheng.
\newblock Repa-e: Unlocking vae for end-to-end tuning of latent diffusion transformers.
\newblock In \emph{CVPR}, 2025.

\bibitem[Levine(2018)]{levine2018reinforcement}
Sergey Levine.
\newblock Reinforcement learning and control as probabilistic inference: Tutorial and review.
\newblock \emph{arXiv preprint arXiv:1805.00909}, 2018.

\bibitem[Li et~al.(2025)Li, Zhang, Wang, Tian, Tan, Liu, Yu, Xie, Lu, Wang, and Lei]{Li2025MergeVQAU}
Siyuan Li, Luyuan Zhang, Zedong Wang, Juanxi Tian, Cheng Tan, Zicheng Liu, Chang Yu, Qingsong Xie, Haonan Lu, Haoqian Wang, and Zhen Lei.
\newblock Mergevq: A unified framework for visual generation and representation with disentangled token merging and quantization.
\newblock In \emph{CVPR}, 2025.

\bibitem[Liao et~al.(2026)Liao, He, Xu, Qu, Li, Wei, and Yao]{liao2025va}
Xinyao Liao, Qiyuan He, Kai Xu, Xiaoye Qu, Yicong Li, Wei Wei, and Angela Yao.
\newblock Va-$\pi$: Variational policy alignment for pixel-aware autoregressive generation.
\newblock In \emph{CVPR}, 2026.

\bibitem[Lin et~al.(2014)Lin, Maire, Belongie, Hays, Perona, Ramanan, Doll{\'a}r, and Zitnick]{lin2014microsoft}
Tsung-Yi Lin, Michael Maire, Serge Belongie, James Hays, Pietro Perona, Deva Ramanan, Piotr Doll{\'a}r, and C~Lawrence Zitnick.
\newblock Microsoft {COCO}: Common objects in context.
\newblock In \emph{ECCV}, 2014.

\bibitem[Lipman et~al.(2023)Lipman, Chen, Ben-Hamu, Nickel, and Le]{lipman2022flow}
Yaron Lipman, Ricky~TQ Chen, Heli Ben-Hamu, Maximilian Nickel, and Matt Le.
\newblock Flow matching for generative modeling.
\newblock In \emph{ICLR}, 2023.

\bibitem[Liu et~al.(2025)Liu, Liu, Liang, Li, Liu, Wang, Wan, Zhang, and Ouyang]{Liu2025FlowGRPO}
Jie Liu, Gongye Liu, Jiajun Liang, Yangguang Li, Jiaheng Liu, Xintao Wang, Pengfei Wan, Di~Zhang, and Wanli Ouyang.
\newblock Flow-grpo: Training flow matching models via online rl.
\newblock In \emph{NeurIPS}, 2025.

\bibitem[Loshchilov and Hutter(2019)]{loshchilov2017decoupled}
Ilya Loshchilov and Frank Hutter.
\newblock Decoupled weight decay regularization.
\newblock In \emph{ICLR}, 2019.

\bibitem[Luo et~al.(2024)Luo, Shi, Ge, Yang, Wang, and Shan]{Luo2024OpenMAGVIT2AO}
Zhuoyan Luo, Fengyuan Shi, Yixiao Ge, Yujiu Yang, Limin Wang, and Ying Shan.
\newblock Open-magvit2: An open-source project toward democratizing auto-regressive visual generation.
\newblock \emph{arXiv preprint arXiv:2409.04410}, 2024.

\bibitem[Neal and Hinton(1998)]{neal1998view}
Radford~M Neal and Geoffrey~E Hinton.
\newblock A view of the em algorithm that justifies incremental, sparse, and other variants.
\newblock In \emph{Learning in graphical models}, pages 355--368. Springer, 1998.

\bibitem[Ouyang et~al.(2022)Ouyang, Wu, Jiang, Almeida, Wainwright, Mishkin, Zhang, Agarwal, Slama, Ray, et~al.]{ouyang2022training}
Long Ouyang, Jeffrey Wu, Xu~Jiang, Diogo Almeida, Carroll Wainwright, Pamela Mishkin, Chong Zhang, Sandhini Agarwal, Katarina Slama, Alex Ray, et~al.
\newblock Training language models to follow instructions with human feedback.
\newblock \emph{NeurIPS}, 2022.

\bibitem[Pang et~al.(2024)Pang, Zhang, Luan, Man, Tan, Zhang, Freeman, and Wang]{Pang2024RandARDA}
Ziqi Pang, Tianyuan Zhang, Fujun Luan, Yunze Man, Hao Tan, Kai Zhang, William~T. Freeman, and Yu-Xiong Wang.
\newblock Randar: Decoder-only autoregressive visual generation in random orders.
\newblock In \emph{CVPR}, 2024.

\bibitem[Peebles and Xie(2023)]{peebles2023dit}
William Peebles and Saining Xie.
\newblock Scalable diffusion models with transformers.
\newblock In \emph{ICCV}, 2023.

\bibitem[Peng et~al.(2019)Peng, Kumar, Zhang, and Levine]{peng2019advantage}
Xue~Bin Peng, Aravind Kumar, Grace Zhang, and Sergey Levine.
\newblock Advantage-weighted regression: Simple and scalable off-policy reinforcement learning.
\newblock \emph{arXiv preprint arXiv:1910.00177}, 2019.

\bibitem[Peters and Schaal(2007)]{peters2007reinforcement}
Jan Peters and Stefan Schaal.
\newblock Reinforcement learning by reward-weighted regression for operational space control.
\newblock In \emph{ICML}, 2007.

\bibitem[Podell et~al.(2024)Podell, English, Lacey, Blattmann, Dockhorn, M{\"u}ller, Penna, and Rombach]{Podell2023SDXLIL}
Dustin Podell, Zion English, Kyle Lacey, Andreas Blattmann, Tim Dockhorn, Jonas M{\"u}ller, Joe Penna, and Robin Rombach.
\newblock Sdxl: Improving latent diffusion models for high-resolution image synthesis.
\newblock In \emph{ICLR}, 2024.

\bibitem[Prabhudesai et~al.(2023)Prabhudesai, Goyal, Pathak, and Fragkiadaki]{prabhudesai2023aligning}
Mihir Prabhudesai, Anirudh Goyal, Deepak Pathak, and Katerina Fragkiadaki.
\newblock Aligning text-to-image diffusion models with reward backpropagation.
\newblock \emph{arXiv preprint arXiv:2310.03739}, 2023.

\bibitem[{Qwen Team}(2024)]{qwen2024qwen25}
{Qwen Team}.
\newblock Qwen2.5 technical report.
\newblock \emph{arXiv preprint}, 2024.

\bibitem[Radford et~al.(2021)Radford, Kim, Hallacy, Ramesh, Goh, Agarwal, Sastry, Askell, Mishkin, Clark, et~al.]{radford2021learning}
Alec Radford, Jong~Wook Kim, Chris Hallacy, Aditya Ramesh, Gabriel Goh, Sandhini Agarwal, Girish Sastry, Amanda Askell, Pamela Mishkin, Jack Clark, et~al.
\newblock Learning transferable visual models from natural language supervision.
\newblock In \emph{ICML}, 2021.

\bibitem[Ramesh et~al.(2021)Ramesh, Pavlov, Goh, Gray, Voss, Radford, Chen, and Sutskever]{ramesh2021zero}
Aditya Ramesh, Mikhail Pavlov, Gabriel Goh, Scott Gray, Chelsea Voss, Alec Radford, Mark Chen, and Ilya Sutskever.
\newblock Zero-shot text-to-image generation.
\newblock In \emph{ICML}, 2021.

\bibitem[Ranzato et~al.(2016)Ranzato, Chopra, Auli, and Zaremba]{ranzato2015sequence}
Marc'Aurelio Ranzato, Sumit Chopra, Michael Auli, and Wojciech Zaremba.
\newblock Sequence level training with recurrent neural networks.
\newblock In \emph{ICLR}, 2016.

\bibitem[Razavi et~al.(2019)Razavi, Van~den Oord, and Vinyals]{razavi2019generating}
Ali Razavi, Aaron Van~den Oord, and Oriol Vinyals.
\newblock Generating diverse high-fidelity images with {VQ-VAE-2}.
\newblock In \emph{NeurIPS}, 2019.

\bibitem[Rombach et~al.(2022)Rombach, Blattmann, Lorenz, Esser, and Ommer]{rombach2022high}
Robin Rombach, Andreas Blattmann, Dominik Lorenz, Patrick Esser, and Bj{\"o}rn Ommer.
\newblock High-resolution image synthesis with latent diffusion models.
\newblock In \emph{CVPR}, 2022.

\bibitem[Schulman et~al.(2017)Schulman, Wolski, Dhariwal, Radford, and Klimov]{schulman2017proximal}
John Schulman, Filip Wolski, Prafulla Dhariwal, Alec Radford, and Oleg Klimov.
\newblock Proximal policy optimization algorithms.
\newblock In \emph{arXiv preprint arXiv:1707.06347}, 2017.

\bibitem[Shao et~al.(2024)Shao, Wang, Zhu, Xu, Song, Zhang, Li, Wu, and Guo]{shao2024deepseekmath}
Zhihong Shao, Peiyi Wang, Qihao Zhu, Runxin Xu, Junxiao Song, Mingchuan Zhang, YK~Li, Y~Wu, and Daya Guo.
\newblock {DeepSeekMath}: Pushing the limits of mathematical reasoning in open language models.
\newblock \emph{arXiv preprint arXiv:2402.03300}, 2024.

\bibitem[Shi et~al.(2025)Shi, Luo, Ge, Yang, Shan, and Wang]{Shi2024ScalableIT}
Fengyuan Shi, Zhuoyan Luo, Yixiao Ge, Yujiu Yang, Ying Shan, and Limin Wang.
\newblock Scalable image tokenization with index backpropagation quantization.
\newblock In \emph{CVPR}, 2025.

\bibitem[Sun et~al.(2023)Sun, Pan, Ge, Li, Duan, Wu, Zhang, Zhou, Qin, Wang, et~al.]{sun2023journeydb}
Keqiang Sun, Junting Pan, Yuying Ge, Hao Li, Haodong Duan, Xiaoshi Wu, Renrui Zhang, Aojun Zhou, Zipeng Qin, Yi~Wang, et~al.
\newblock Journeydb: A benchmark for generative image understanding.
\newblock \emph{NeurIPS}, 2023.

\bibitem[Sun et~al.(2024)Sun, Jiang, Chen, Zhang, Peng, Luo, and Yuan]{sun2024llamagen}
Peize Sun, Yi~Jiang, Shoufa Chen, Shilong Zhang, Bingyue Peng, Ping Luo, and Zehuan Yuan.
\newblock Autoregressive model beats diffusion: {L}lama for scalable image generation.
\newblock \emph{arXiv preprint arXiv:2406.06525}, 2024.

\bibitem[Tarvainen and Valpola(2017)]{tarvainen2017mean}
Antti Tarvainen and Harri Valpola.
\newblock Mean teachers are better role models: Weight-averaged consistency targets improve semi-supervised learning results.
\newblock In \emph{NeurIPS}, 2017.

\bibitem[Tian et~al.(2024)Tian, Jiang, Yuan, Peng, and Wang]{tian2024visual}
Keyu Tian, Yi~Jiang, Zehuan Yuan, Bingyue Peng, and Liwei Wang.
\newblock Visual autoregressive modeling: Scalable image generation via next-scale prediction.
\newblock \emph{NeurIPS}, 2024.

\bibitem[Van Den~Oord et~al.(2017)Van Den~Oord, Vinyals, et~al.]{van2017neural}
Aaron Van Den~Oord, Oriol Vinyals, et~al.
\newblock Neural discrete representation learning.
\newblock In \emph{NeurIPS}, 2017.

\bibitem[Wallace et~al.(2024)Wallace, Dang, Rafailov, Zhou, Lou, Purushwalkam, Ermon, Xiong, Joty, and Naik]{wallace2023diffusion}
Bram Wallace, Meihua Dang, Rafael Rafailov, Linqi Zhou, Aaron Lou, Senthil Purushwalkam, Stefano Ermon, Caiming Xiong, Shafiq Joty, and Nikhil Naik.
\newblock Diffusion model alignment using direct preference optimization.
\newblock In \emph{CVPR}, 2024.

\bibitem[Wang et~al.(2025)Wang, Tian, Wang, Zhang, Huang, Wu, and Jiang]{simplear2024}
Junke Wang, Zhi Tian, Xun Wang, Xinyu Zhang, Weilin Huang, Zuxuan Wu, and Yu-Gang Jiang.
\newblock Simplear: Pushing the frontier of autoregressive visual generation through pretraining, sft, and rl.
\newblock \emph{arXiv preprint arXiv:2504.11455}, 2025.

\bibitem[Wang et~al.(2024)Wang, Zhang, Luo, Sun, Cui, Wang, Zhang, Wang, Li, Yu, et~al.]{wang2024emu3}
Xinlong Wang, Xiaosong Zhang, Zhengxiong Luo, Quan Sun, Yufeng Cui, Jinsheng Wang, Fan Zhang, Yueze Wang, Zhen Li, Qiying Yu, et~al.
\newblock Emu3: Next-token prediction is all you need.
\newblock \emph{arXiv preprint arXiv:2409.18869}, 2024.

\bibitem[Williams(1992)]{williams1992simple}
Ronald~J. Williams.
\newblock Simple statistical gradient-following algorithms for connectionist reinforcement learning.
\newblock \emph{Machine Learning}, 1992.

\bibitem[Wu(1983)]{wu1983convergence}
C.~F.~Jeff Wu.
\newblock On the convergence properties of the {EM} algorithm.
\newblock \emph{The Annals of Statistics}, 11\penalty0 (1):\penalty0 95--103, 1983.

\bibitem[Wu et~al.(2023)Wu, Hao, Sun, Chen, Zhu, Zhao, and Li]{wu2023hpsv2}
Xiaoshi Wu, Yiming Hao, Keqiang Sun, Yixiong Chen, Feng Zhu, Rui Zhao, and Hongsheng Li.
\newblock Human preference score v2: A solid benchmark for evaluating human preferences of text-to-image synthesis.
\newblock \emph{arXiv preprint arXiv:2306.09341}, 2023.

\bibitem[Xiong et~al.(2025)Xiong, Liew, Huang, Feng, and Liu]{Xiong2025GigaTokSV}
Tianwei Xiong, Jun~Hao Liew, Zilong Huang, Jiashi Feng, and Xihui Liu.
\newblock Gigatok: Scaling visual tokenizers to 3 billion parameters for autoregressive image generation.
\newblock In \emph{CVPR}, 2025.

\bibitem[Xu et~al.(2023)Xu, Liu, Wu, Tong, Li, Ding, Tang, and Dong]{xu2024imagereward}
Jiazheng Xu, Xiao Liu, Yuchen Wu, Yuxuan Tong, Qinkai Li, Ming Ding, Jie Tang, and Yuxiao Dong.
\newblock Imagereward: Learning and evaluating human preferences for text-to-image generation.
\newblock In \emph{NeurIPS}, 2023.

\bibitem[Yu et~al.(2022)Yu, Xu, Koh, Luong, Baid, Wang, Vasudevan, Ku, Yang, Amin~Karbasi, et~al.]{yu2022parti}
Jiahui Yu, Yuanzhong Xu, Jing~Yu Koh, Thang Luong, Gunjan Baid, Zirui Wang, Vijay Vasudevan, Alexander Ku, Yinfei Yang, Burcu Amin~Karbasi, et~al.
\newblock Scaling autoregressive models for content-rich text-to-image generation.
\newblock \emph{TMLR}, 2022.

\bibitem[Yu et~al.(2024)Yu, Weber, Deng, Shen, Cremers, and Chen]{Yu2024AnII}
Qihang Yu, Mark Weber, Xueqing Deng, Xiaohui Shen, Daniel Cremers, and Liang-Chieh Chen.
\newblock An image is worth 32 tokens for reconstruction and generation.
\newblock In \emph{NeurIPS}, 2024.

\bibitem[Zhang et~al.(2026)Zhang, Yu, Ma, Zhang, Pan, Yao, Xiao, Huang, and Zhao]{zhang2026group}
Guohui Zhang, Hu~Yu, Xiaoxiao Ma, Jinghao Zhang, Yaning Pan, Mingde Yao, Jie Xiao, Linjiang Huang, and Feng Zhao.
\newblock Group critical-token policy optimization for autoregressive image generation.
\newblock In \emph{ICLR}, 2026.

\bibitem[Zhang et~al.(2018)Zhang, Isola, Efros, Shechtman, and Wang]{zhang2018unreasonable}
Richard Zhang, Phillip Isola, Alexei~A Efros, Eli Shechtman, and Oliver Wang.
\newblock The unreasonable effectiveness of deep features as a perceptual metric.
\newblock In \emph{CVPR}, 2018.

\bibitem[Zhou et~al.(2024)Zhou, Yu, Babu, Tirumala, Yasunaga, Shamis, Kahn, Ma, Zettlemoyer, and Levy]{Zhou2024Transfusion}
Chunting Zhou, Lili Yu, Arun Babu, Kushal Tirumala, Michihiro Yasunaga, Leonid Shamis, Jacob Kahn, Xuezhe Ma, Luke Zettlemoyer, and Omer Levy.
\newblock Transfusion: Predict the next token and diffuse images with one multi-modal model.
\newblock \emph{arXiv preprint arXiv:2408.11039}, 2024.

\end{thebibliography}
}

\newpage

\appendix
\appendix
\numberwithin{table}{section}
\numberwithin{figure}{section}
\renewcommand{\thetable}{\thesection.\arabic{table}}
\setcounter{table}{0}
\renewcommand{\thefigure}{\thesection.\arabic{figure}}
\setcounter{figure}{0}

\clearpage
\appendix
\begin{center}
{\LARGE\bfseries Appendix for RankE}
\end{center}
\vspace{0.8em}

\section*{Roadmap}
The appendix is organized into three parts, progressing from theoretical grounding through empirical validation to implementation details. %

\smallskip
\noindent\textbf{Part~I -- Theoretical Foundations} (App.~\ref{app:related_extended}--\ref{app:em}) positions RankE in the literature and formalizes its GEM convergence guarantee, answering \emph{why alternating co-evolution is principled, not heuristic.} %

\smallskip
\noindent\textbf{Part~II -- Mechanism Diagnostics and Robustness} (App.~\ref{app:covariate_shift}--\ref{app:extended_ablation}) measures Latent Covariate Shift, traces training dynamics across reward signals, and consolidates sensitivity studies, answering \emph{whether RankE works for the reasons claimed and is robust to design choices.} %

\smallskip
\noindent\textbf{Part~III -- Reproducibility} (App.~\ref{app:impl}--\ref{app:data}) records the training algorithm, hyperparameters, compute footprint, and data pipeline, providing everything needed to reproduce RankE end-to-end. %

\vspace{1.2em}
\begin{center}
\rule{0.85\linewidth}{0.4pt}\\[0.4em]
{\large\bfseries Part I\;\;|\;\;Theoretical Foundations}\\[0.25em]
{\itshape\small Where RankE sits in the literature, and why its
alternation is principled rather than ad hoc.}\\[0.3em]
\rule{0.85\linewidth}{0.4pt}
\end{center}
\vspace{0.4em}

\section{Extended Related Work}
\label{app:related_extended}

This appendix expands §\ref{sec:related_work} along three axes that together motivate RankE's alternating design: discrete visual tokenizers, AR generator factorizations, and the gradient barrier between policy and pixels. %

\subsection{Discrete Visual Tokenizers}

Discrete visual tokenizers compress images into compact grids of latent codes that downstream AR models treat as token sequences. %
Most adopt vector quantization (VQ) to learn a finite codebook mapping encoder features to discrete indices, as in VQ-VAE~\citep{van2017neural} and VQGAN~\citep{esser2021taming}. %
Recent tokenizer research follows three complementary directions: (i) improving the quantization mechanism to reduce VQ-induced fidelity loss~\citep{Luo2024OpenMAGVIT2AO,Shi2024ScalableIT}; (ii) adding regularization or representation constraints for more structured tokens that better support downstream generation~\citep{Xiong2025GigaTokSV,Chen2024SoftVQVAEE1}; and (iii) shortening token length via stronger compression for efficiency~\citep{Li2025MergeVQAU,Yu2024AnII}. %
All of these works share one assumption with which RankE explicitly breaks: the decoder, once pre-trained, is treated as a fixed renderer. %

\subsection{AR-based Image Generators}

AR-based image generators model images as token sequences under a chosen factorization. %
Early systems use raster-order decoding with Transformer generators, where scaling yields LLM-like gains~\citep{kaplan2020scaling}, exemplified by DALL-E~\citep{ramesh2021zero}, CogView~\citep{ding2021cogview}, and Parti~\citep{yu2022parti}. %
Subsequent work explores alternative factorizations: next-scale prediction in VAR~\citep{tian2024visual}, masked parallel decoding in MaskGIT~\citep{chang2022maskgit} and MUSE~\citep{chang2023muse}, and randomized orders in RandAR~\citep{Pang2024RandARDA}. %
Despite different decoding schedules, these methods share a two-stage pipeline with a frozen tokenizer: training conditions on ground-truth codes via teacher forcing, while inference conditions on self-sampled codes. %
As §\ref{sec:method} shows, this decoupling becomes a first-order bottleneck \emph{after} preference optimization, motivating our co-trainable decoder. %

\subsection{The Gradient Barrier in Discrete AR}

Direct reward back-propagation does not transfer from diffusion to discrete AR because two non-differentiable operations sit between policy and pixels: categorical sampling~\citep{jang2016categorical} and VQ argmax~\citep{van2017neural}. %
The straight-through estimator (STE)~\citep{bengio2013estimating} is a biased surrogate empirically unstable at the codebook scales ($\sim$16K entries) of modern visual tokenizers~\citep{kaiser2018fast,huh2023straightening}; Gumbel-Softmax~\citep{jang2016categorical} requires temperature annealing and degrades at high vocabulary sizes. %
While REPA-E~\citep{leng2025repa} unlocks the VAE for diffusion via reparameterization through a smooth chain, the discrete AR setting requires a categorically different solution: rather than differentiating \emph{through} the discrete bottleneck, RankE alternates \emph{around} it, updating the policy with a discrete-friendly RL objective and the decoder with continuous gradients on policy-sampled latents. %

\section{Generalized EM Formal Derivation}
\label{app:em}

This appendix formalizes the GEM interpretation of RankE introduced in §\ref{subsec:coevolution}: both stages are stochastic ascent steps on the same MAP-augmented evidence lower bound. %
We further clarify the precise role of advantage normalization, PPO clipping, the Rank-GAN surrogate, and the regularization terms. %

\subsection{Setup and the Optimality Variable}
\label{app:em-setup}

Recall the joint objective in Eq.~\ref{eq:e2e_obj}: $J(\theta, \phi) = \mathbb{E}_{y, z\sim\pi_\theta}[r(D_\phi(z), y)]$. %
Following the control-as-inference framework~\citep{levine2018reinforcement}, we introduce a binary optimality variable $\mathcal{O}$ with conditional likelihood
\begin{equation}
    p(\mathcal{O}{=}1 \mid z, y; \phi) \;=\; \frac{1}{Z(y; \phi)}\,\exp\!\big(r(D_\phi(z), y) / \beta\big),
    \label{eq:opt-app}
\end{equation}
where $\beta$ is the KL coefficient in Eq.~\ref{eq:policy}, $Z(y; \phi) = \mathbb{E}_{z\sim\pi_{\mathrm{ref}}}[\exp(r/\beta)]$ is the prompt-dependent normalizer, and $r$ is upper-bounded so $Z(y;\phi)<\infty$. %
The marginal $\log p(\mathcal{O}{=}1\mid y;\phi)=\log Z(y;\phi)$ is the quantity whose maximization in $\phi$ corresponds to producing high-reward outputs under the prior $\pi_{\mathrm{ref}}$. %

\subsection{The MAP-Augmented ELBO}
\label{app:em-elbo}

Treating $\pi_\theta(\cdot\mid y)$ as a variational posterior and $\pi_{\mathrm{ref}}(\cdot\mid y)$ as the prior, Jensen's inequality yields
\begin{equation}
    \log p(\mathcal{O}{=}1 \mid y; \phi) \;\geq\;
    \underbrace{\frac{1}{\beta}\,\mathbb{E}_{z\sim\pi_\theta}\!\big[r(D_\phi(z), y)\big]
    \;-\;
    D_{\mathrm{KL}}\!\big(\pi_\theta(\cdot\mid y)\,\big\|\,\pi_{\mathrm{ref}}(\cdot\mid y)\big)}_{=:\,\mathcal{F}(\theta, \phi;\,y)}
    + \text{const}.
    \label{eq:elbo-app}
\end{equation}
Adding a $\phi$-prior $\log p(\phi) = -\lambda_r \mathcal{L}_{\mathrm{recon}}(\phi) - \lambda_c \mathcal{L}_{\mathrm{consist}}(\phi) + \text{const}$ that anchors the decoder to its pre-training manifold gives the MAP-augmented bound
\begin{equation}
    \mathcal{L}(\theta, \phi) \;:=\; \mathbb{E}_{y\sim\mathcal{D}}\!\big[\mathcal{F}(\theta, \phi;\,y)\big] + \log p(\phi).
    \label{eq:map-elbo}
\end{equation}
RankE optimizes $\mathcal{L}$ by alternating block-coordinate ascent on $\theta$ (E-step) and $\phi$ (M-step). %

\subsection{E-step: GRPO as ELBO Ascent on \texorpdfstring{$\theta$}{theta}}
\label{app:em-e-step}

Fixing $\phi$, the exact gradient is
\begin{equation}
\nabla_\theta \mathcal{L}
=\frac{1}{\beta}\,\mathbb{E}_{y}\!\Big[\mathbb{E}_{\pi_\theta}\!\big[r\,\nabla_\theta\log\pi_\theta(z\mid y)\big]\Big]
\;-\;\nabla_\theta D_{\mathrm{KL}}(\pi_\theta\,\|\,\pi_{\mathrm{ref}}).
\label{eq:dL-dtheta}
\end{equation}
GRPO replaces $r$ with a group-normalized advantage and clips importance ratios. %
Both modifications are well-known operations on top of the unbiased ELBO gradient. %

\paragraph{(i) Advantage normalization is variance reduction.}
The advantage $A_i = (r_i - \mu_r)/\sigma_r$ subtracts a per-prompt baseline $\mu_r$. %
Subtracting any $y$-dependent baseline $b(y)$ leaves the gradient unchanged in expectation~\citep{williams1992simple}: $\mathbb{E}_{\pi_\theta}[b(y)\nabla_\theta \log \pi_\theta] = b(y)\nabla_\theta \mathbb{E}_{\pi_\theta}[1] = 0$. %
Dividing by $\sigma_r$ rescales the gradient per prompt, preserving the ascent direction. %
Hence advantage normalization is an unbiased, variance-reduced estimator of the reward gradient. %

\paragraph{(ii) PPO clipping is trust-region projection.}
The clipped objective $\min\!\big(\rho_i A_i,\;\mathrm{clip}(\rho_i,1{\pm}\epsilon)A_i\big)$ projects the update onto a trust region in importance-ratio space~\citep{schulman2017proximal}. %
Within $|\rho_i - 1|\leq\epsilon$ the clipped objective coincides with the unclipped one; outside, the gradient is zeroed for samples that would push the policy too far in one step. %
Clipping introduces finite-step bias of order $\epsilon$; for our $\epsilon{=}0.2$ the expected ascent direction matches the unclipped gradient. %

\paragraph{(iii) The KL term directly implements ELBO regularization.}
The penalty $\beta D_{\mathrm{KL}}(\pi_\theta\|\pi_{\mathrm{ref}})$ in Eq.~\ref{eq:policy} is precisely the second term of Eq.~\ref{eq:dL-dtheta}, with the EMA reference $\pi_{\mathrm{ref}}$ acting as the slowly-varying latent prior---the standard incremental EM regime~\citep{neal1998view}. %

\paragraph{Summary.}
Combining (i)--(iii), the GRPO update is a stochastic, variance-reduced, trust-region-clipped ascent step on $\nabla_\theta \mathcal{L}$. %

\subsection{M-step: Decoder Update as MAP Ascent on \texorpdfstring{$\phi$}{phi}}
\label{app:em-m-step}

Fixing $\theta$, the gradient w.r.t.\ $\phi$ is
\begin{equation}
\nabla_\phi \mathcal{L}
=\frac{1}{\beta}\,\mathbb{E}_{y}\!\Big[\mathbb{E}_{\pi_\theta}\!\big[\nabla_\phi r(D_\phi(z), y)\big]\Big]
\;+\;\nabla_\phi \log p(\phi),
\label{eq:dL-dphi}
\end{equation}
since the KL term in $\mathcal{F}$ does not depend on $\phi$. %

\paragraph{(i) Differentiable rewards.}
When $r$ is differentiable in $\phi$ (e.g., CLIP), the first term in Eq.~\ref{eq:dL-dphi} is exactly $-\nabla_\phi \mathcal{L}_{\mathrm{reward}}$, with $1/\beta$ absorbed into $\lambda_d$. %

\paragraph{(ii) Non-differentiable rewards: Rank-GAN as importance-weighted surrogate.}
When $r$ is a black-box (e.g., HPSv2), $\nabla_\phi r$ is unavailable. %
The reward-tilted optimal posterior is $p^*(z\mid y) \propto \pi_{\mathrm{ref}}(z\mid y)\exp(r/\beta)$; at the optimum $\pi_\theta\approx p^*$ the importance weight from $\pi_\theta$ to $p^*$ simplifies to $w(z) \propto \exp(r/\beta)$, recovering our $w(\hat{z}_i)\propto\exp(r_i/\tau)$ with $\tau$ as a softening temperature~\citep{peters2007reinforcement,peng2019advantage}. %
The Rank-GAN loss
\begin{equation}
    \mathcal{L}_{\mathrm{Rank\text{-}GAN}}(\phi)
    = -\,\mathbb{E}_{\hat z\sim\pi_\theta}\!\big[\,w(\hat z)\,\mathrm{Disc}(D_\phi(\hat z))\,\big]
    \label{eq:rankgan-app}
\end{equation}
applies the standard GAN objective~\citep{goodfellow2014generative} to a reward-tilted distribution: it pushes $D_\phi$ to make high-reward decoded samples indistinguishable from real images, the GAN density-ratio analogue of $\mathbb{E}_{p^*}[\log p_\phi(x\mid z)]$. %

\paragraph{(iii) The prior gradient.}
$\nabla_\phi \log p(\phi) = -\lambda_r \nabla_\phi \mathcal{L}_{\mathrm{recon}} - \lambda_c \nabla_\phi \mathcal{L}_{\mathrm{consist}}$, exactly the regularization components of Eq.~\ref{eq:decoder}. %

\paragraph{Summary.}
The decoder update is stochastic ascent on $\nabla_\phi \mathcal{L}$, with $\mathcal{L}_{\mathrm{Rank\text{-}GAN}}$ as a principled surrogate when the reward is non-differentiable. %

\subsection{Convergence}
\label{app:em-convergence}

Both stages perform stochastic ascent on the same $\mathcal{L}$. %
Under standard Robbins--Monro learning-rate conditions and bounded gradient variance, the iteration converges to a stationary point of $\mathcal{L}$~\citep{neal1998view}. %
This is the GEM guarantee: each stage need only \emph{improve} $\mathcal{L}$ given the other---generalizing classical EM~\citep{dempster1977maximum,wu1983convergence}. %

\vspace{1.2em}
\begin{center}
\rule{0.85\linewidth}{0.4pt}\\[0.4em]
{\large\bfseries Part II\;\;|\;\;Mechanism Diagnostics and Robustness}\\[0.25em]
{\itshape\small Empirical evidence that RankE's gains come from
the mechanism it was designed to engage, and that the design is
robust within reasonable ranges.}\\[0.3em]
\rule{0.85\linewidth}{0.4pt}
\end{center}
\vspace{0.4em}

\section{Latent Covariate Shift: Measurement Protocol}
\label{app:covariate_shift}

This appendix details how the Latent Covariate Shift diagnostic reported in Fig.~\ref{fig:combined} and Fig.~\ref{fig:analyse_combine_shift_entropy} is computed, so that the curves are reproducible from token statistics alone. %

\paragraph{Token extraction protocol.}
\textit{Ground-truth tokens.} We sample $N{=}5{,}000$ images from MS-COCO 2014 validation~\citep{lin2014microsoft} and encode via the frozen VQ tokenizer: $\mathbf{z}=\arg\min_{c\in\mathcal{C}}\|\mathrm{Enc}(x)_{ij}-c\|_2$. %
\textit{Policy-sampled tokens.} For each checkpoint $\pi_\theta^{(t)}$ at $t\in\{500,1000,\ldots,6000\}$, we generate $N{=}5{,}000$ sequences from the same COCO prompts using nucleus sampling ($p{=}0.9$, temperature $1.0$, top-$k{=}1000$, CFG $6.0$, fixed seed). %
We record raw token indices without pixel decoding to isolate the latent distribution from decoder artifacts. %

\paragraph{Divergence metric.}
Let $p_\mathrm{GT}(c)$ and $p_\theta(c)$ denote the empirical unigram frequency of codebook entry $c$. %
We compute
\[
D_\mathrm{KL}(p_\theta\|p_\mathrm{GT})
  = \sum_{c} p_\theta(c)\log\frac{p_\theta(c)}{p_\mathrm{GT}(c)}.
\]
A monotonically increasing $D_\mathrm{KL}$ over training quantifies the worsening covariate shift, directly motivating decoder adaptation in RankE. %

\section{Training Dynamics and Convergence Behavior}
\label{app:dynamics}

Having defined the diagnostic, we now ask how the loss components themselves behave during co-evolution. %
This appendix reports per-step dynamics under both reward signals (§\ref{app:dynamics-perstep}) and convergence patterns across co-evolution rounds (§\ref{app:convergence}). %

\subsection{Per-Step Dynamics under CLIP and HPSv2}
\label{app:dynamics-perstep}

\begin{figure}[t]
    \centering
    \includegraphics[width=\textwidth]{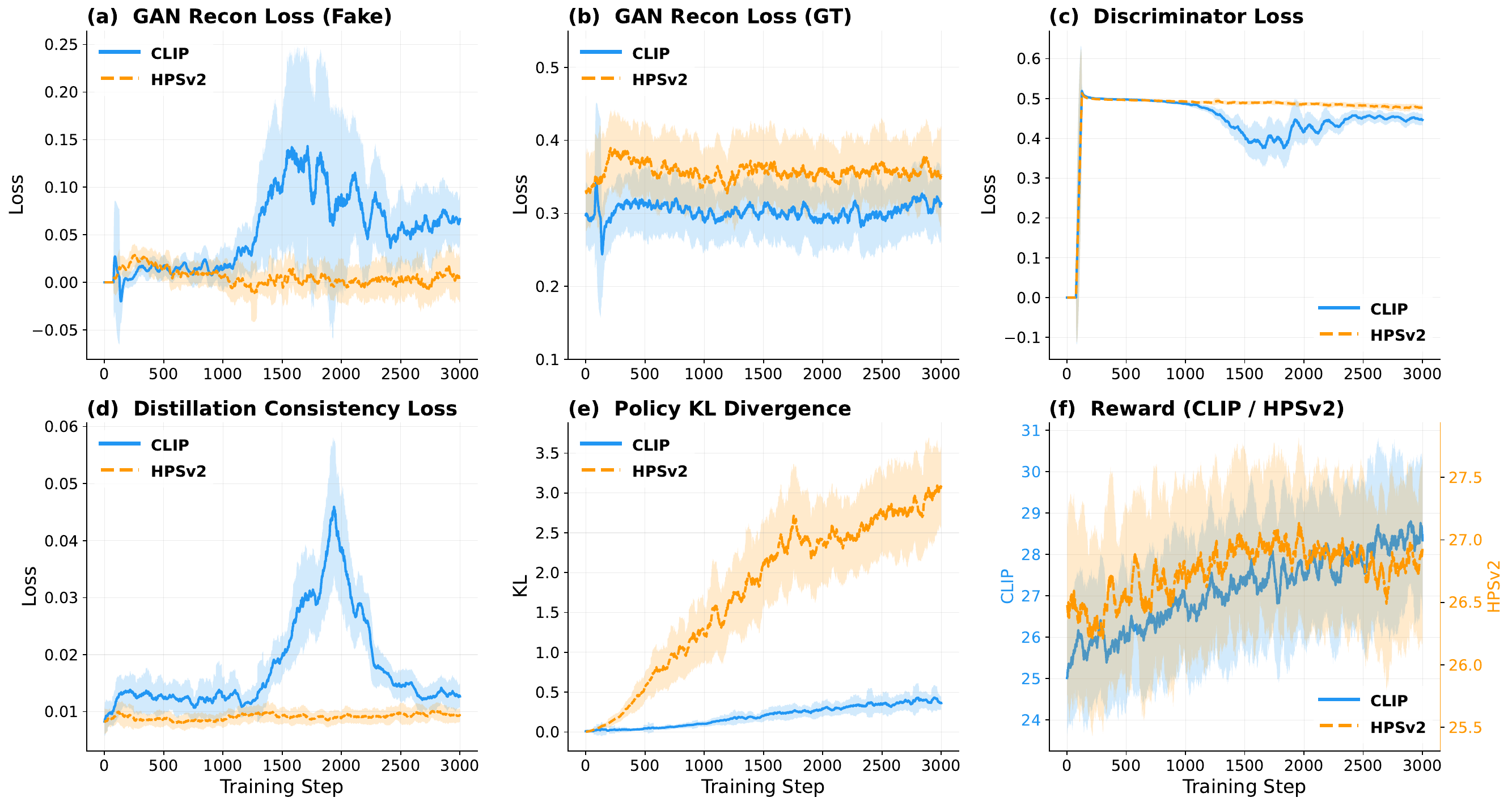}
    \caption{Training dynamics of RankE over $3{,}000$ steps under
    \textcolor[HTML]{2196F3}{\textbf{CLIP}} (blue, solid) and
    \textcolor[HTML]{FF9800}{\textbf{HPSv2}} (orange, dashed).
    (a, b) GAN reconstruction loss on generated and ground-truth
    images. (c, d) Discriminator and distillation consistency
    losses. (e) Policy KL divergence. (f) Reward curves with dual
    $y$-axes.}
    \label{fig:training_dynamics_merged}
\end{figure}

Fig.~\ref{fig:training_dynamics_merged} tracks six diagnostics over $3{,}000$ steps on LlamaGen-XL with identical hyperparameters under CLIP and HPSv2 rewards. %
The two reward settings produce qualitatively identical convergence behavior across all panels, confirming reward-agnostic generalizability:
\begin{itemize}[leftmargin=*, itemsep=2pt]
    \item \textbf{Decoder fidelity (a, b).} Reconstruction loss on
    policy-sampled images stays near zero (HPSv2) or briefly spikes
    around steps $1{,}000$--$2{,}000$ before recovering (CLIP); GT
    reconstruction stays in $0.3$--$0.4$, with HPSv2 marginally
    higher---a mild trade-off, not destabilization.
    \item \textbf{Adversarial dynamics (c, d).} The discriminator
    reaches $\sim$$0.5$ within 500 steps (declining to $\sim$$0.35$
    under CLIP), and consistency loss stays below $0.015$,
    confirming that EMA-anchored consistency distillation prevents
    abrupt decoder drift.
    \item \textbf{Policy-side dynamics (e, f).} KL rises to
    $2.5$--$3.0$ under both rewards with nearly identical
    trajectories, and rewards rise monotonically
    (CLIP $\sim$21$\to$29, HPSv2 $\sim$25.5$\to$27.5); the steeper
    CLIP curve is consistent with its larger gradient signal in
    (a, d).
\end{itemize}

\subsection{Convergence across Co-Evolution Rounds}
\label{app:convergence}

Across rounds we observe three consistent patterns: rewards rise monotonically; FID transiently worsens at each E-step due to Latent Covariate Shift, then recovers during the subsequent M-step (visible in Fig.~\ref{fig:training_dynamics_merged}(b) and Fig.~\ref{fig:related_teaser}(b) of the main paper); and KL stays bounded under GRPO regularization. %
Together, these confirm the core mechanism: each M-step recalibrates the decoder to the policy's evolving distribution, sustaining a virtuous improvement cycle. %

\section{Extended Ablations and Sensitivity Studies}
\label{app:extended_ablation}

The main ablations isolate \emph{which} components matter; this appendix probes \emph{how sensitive} each component is to its hyperparameter. %
We sweep four design axes on LlamaGen-XL with CLIP reward (MS-COCO 30K), in decreasing order of architectural importance: consistency weight, IS temperature, and EMA decay. %
RankE is robust within reasonable ranges; the only sensitive regimes are sequential ($K{=}1$) training and excessive $\lambda_c$. %

\subsection{Consistency Distillation Weight \texorpdfstring{$\lambda_c$}{lambda c}}
\label{app:consist}

\begin{figure}[t]
    \centering
    \includegraphics[width=\textwidth]{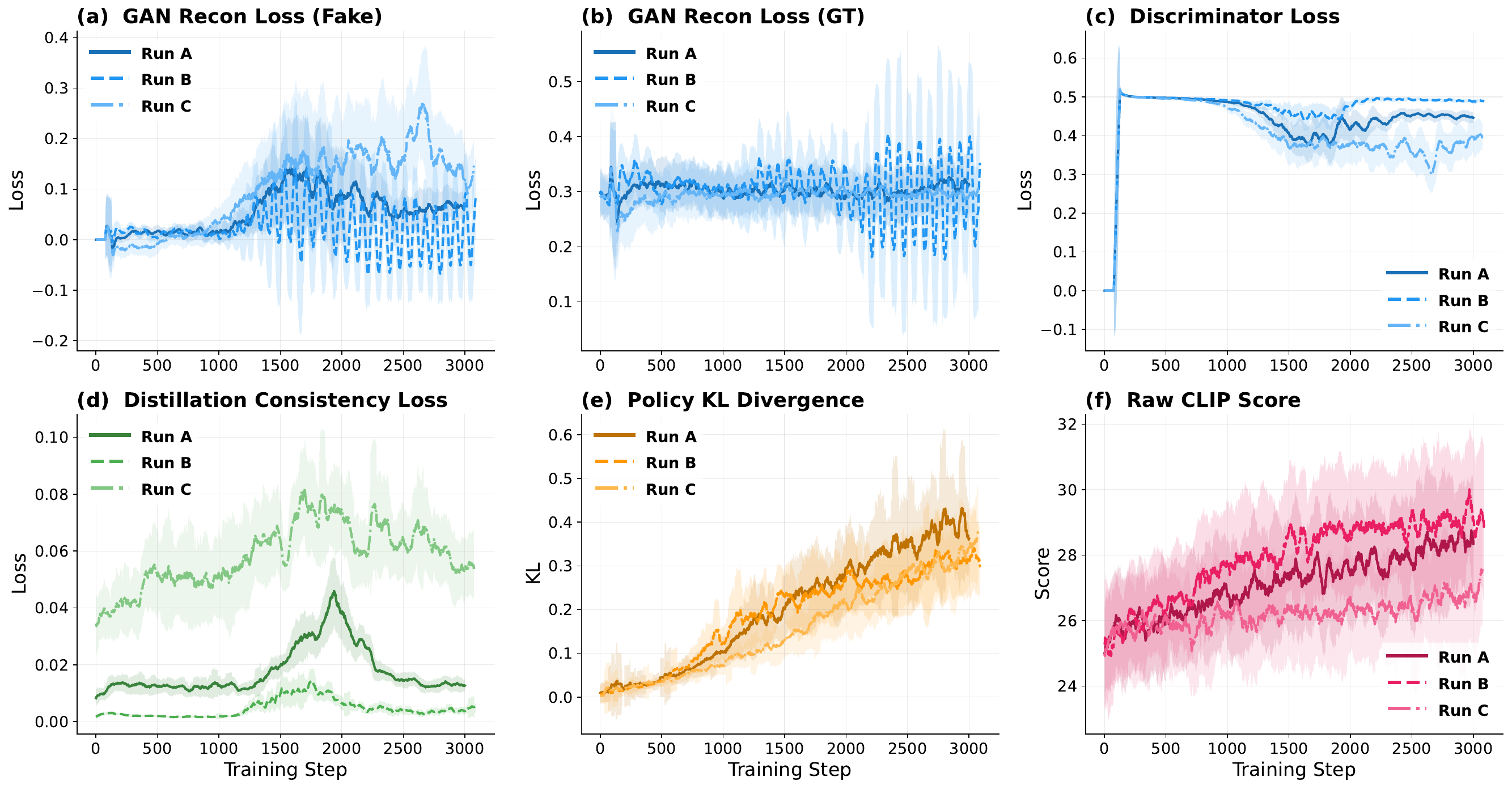}
    \caption{Training dynamics for the $\lambda_c$ ablation.
    Run~A: $\lambda_c{=}10$; Run~B: $\lambda_c{=}1$ (default);
    Run~C: $\lambda_c{=}50$. Excessive $\lambda_c$ causes decoder
    collapse (Run~C); moderate values stay stable.}
    \label{fig:ablation_consist}
\end{figure}

We ablate $\lambda_c$ across $\lambda_c{=}10$ (Run~A), $\lambda_c{=}1$ (Run~B, default), and $\lambda_c{=}50$ (Run~C). %
Dynamics are in Fig.~\ref{fig:ablation_consist}; final metrics in Table~\ref{tab:ablation_consist}. %
Run~B yields the most stable training. %
Run~A introduces mild drift but achieves the best FID/GenEval. %
Run~C causes decoder collapse around step $1{,}500$ (GT reconstruction diverges, discriminator drops to near zero), showing that excessive consistency overwhelms adversarial signal. %
We adopt $\lambda_c{=}1$ for overall stability. %

\begin{table}[ht]
\centering
\small
\caption{Ablation on $\lambda_c$}
\label{tab:ablation_consist}
\renewcommand{\arraystretch}{1.1}
\begin{tabular}{lcccc}
\toprule
\textbf{Setting} & $\lambda_c$ & \textbf{CLIP}\,$\uparrow$ & \textbf{FID}\,$\downarrow$ & \textbf{GenEval}\,$\uparrow$ \\
\midrule
Run~A             & 10 & 33.64          & \textbf{17.37} & \textbf{0.425} \\
Run~B (default)   & 1  & \textbf{34.13} & 18.78          & 0.394          \\
Run~C             & 50 & 32.98          & 17.75          & 0.416          \\
\bottomrule
\end{tabular}
\end{table}

\subsection{Importance Sampling Temperature \texorpdfstring{$\tau$}{tau}}
\label{app:tau_sensitivity}

Table~\ref{tab:tau_sensitivity} evaluates the importance sampling temperature $\tau$. %
A high temperature reduces to uniform sampling, diluting the reward signal and yielding suboptimal results. %
Conversely, an overly low temperature enforces hard selection, which collapses diversity and severely degrades FID to $16.12$. %
The default $\tau{=}0.1$ strikes the optimal balance, achieving the best performance across all three metrics. %

\begin{table}[ht]
\centering
\small
\caption{Sensitivity to importance sampling temperature $\tau$.}
\label{tab:tau_sensitivity}
\renewcommand{\arraystretch}{1.1}
\begin{tabular}{lccc}
\toprule
$\tau$ & FID\,$\downarrow$ & CLIP\,$\uparrow$ & GenEval\,$\uparrow$ \\
\midrule
1.0 (uniform)          & 15.65          & 33.43          & 0.410          \\
\textbf{0.1} (default) & \textbf{15.21} & \textbf{33.76} & \textbf{0.425} \\
0.01 (hard)            & 16.12          & 33.49          & 0.408          \\
\bottomrule
\end{tabular}
\end{table}

\subsection{EMA Decay Rate \texorpdfstring{$\alpha$}{alpha}}
\label{app:ema_decay}

Table~\ref{tab:ema_decay} evaluates the EMA decay rate $\alpha$. %
A fast teacher ($\alpha{=}0.900$) tracks the student too closely, yielding suboptimal fidelity (FID $15.75$) and alignment (CLIP $33.48$). %
Progressively slowing the teacher improves performance, with the default $\alpha{=}0.999$ striking the best stability--adaptability trade-off and achieving optimal results across all three metrics. %
\begin{table}[ht]
\centering
\small
\caption{Effect of EMA decay rate $\alpha$.}
\label{tab:ema_decay}
\renewcommand{\arraystretch}{1.1}
\begin{tabular}{lccc}
\toprule
$\alpha$ & FID\,$\downarrow$ & CLIP\,$\uparrow$ & GenEval\,$\uparrow$ \\
\midrule
0.900                       & 15.75          & 33.48          & 0.412          \\
0.990                      & 15.45          & 33.63          & 0.419          \\
\textbf{0.999} (default)   & \textbf{15.21} & \textbf{33.76} & \textbf{0.425} \\
\bottomrule
\end{tabular}
\end{table}

\vspace{1.2em}
\begin{center}
\rule{0.85\linewidth}{0.4pt}\\[0.4em]
{\large\bfseries Part III\;\;|\;\;Reproducibility}\\[0.25em]
{\itshape\small Everything needed to reproduce RankE end-to-end:
hyperparameters, pseudocode, and data pipeline.}\\[0.3em]
\rule{0.85\linewidth}{0.4pt}
\end{center}
\vspace{0.4em}

\section{Implementation Details and Hyperparameters}
\label{app:impl}

\paragraph{Implementation.}
We implement RankE in PyTorch and train on $8{\times}$NVIDIA A100 GPUs (bf16). %
Stage~1 optimizes the AR policy with GRPO~\citep{shao2024deepseekmath} at learning rate $1{\times}10^{-5}$, with the KL coefficient $\beta$ scheduled against an EMA reference policy to suppress reward exploitation~\citep{gao2023scaling}. %
Stage~2 freezes the policy and updates the decoder with AdamW~\citep{loshchilov2017decoupled} at $5{\times}10^{-5}$, supervised by a PatchGAN discriminator~\citep{isola2017image} and an EMA teacher with decay $0.999$. %

\paragraph{Training hyperparameters.}
Table~\ref{tab:hyperparams_combined} lists the full configurations for both stages. %
\begin{table}[htb]
\centering
\small
\caption{Hyperparameters for Stage~1 (Policy Alignment) and
Stage~2 (Decoder Adaptation).}
\label{tab:hyperparams_combined}
\renewcommand{\arraystretch}{1.1}
\begin{tabular}{lc@{\hspace{1.2cm}}lc}
\toprule
\multicolumn{4}{c}{\textbf{Stage 1: Policy Alignment}} \\
\midrule
\textbf{Hyperparameter} & \textbf{Value} & \textbf{Hyperparameter} & \textbf{Value} \\
\midrule
Optimizer           & AdamW         & Learning rate       & $1\times10^{-5}$ \\
Weight decay        & 0.01          & GRPO group size $G$ & 8                \\
PPO clip $\epsilon$ & 0.2           & KL penalty $\beta$  & 0.05             \\
Sampling temp.      & 1.0           & Top-$p$             & 0.9              \\
Mixed precision     & \texttt{bf16} &                     &                  \\
\midrule
\multicolumn{4}{c}{\textbf{Stage 2: Decoder Adaptation}} \\
\midrule
\textbf{Hyperparameter} & \textbf{Value} & \textbf{Hyperparameter} & \textbf{Value} \\
\midrule
Optimizer              & AdamW         & Learning rate         & $5\times10^{-5}$ \\
Weight decay           & 0.05          & $(\beta_1,\beta_2)$   & (0.5,\,0.9)      \\
Discriminator          & PatchGAN      & Discriminator lr      & $5\times10^{-5}$ \\
EMA decay $\alpha$     & 0.999         & IS temperature $\tau$ & 0.1              \\
$\lambda_{\mathrm{r}}$ & 1.0           & $\lambda_{\mathrm{g}}$ & 0.5             \\
$\lambda_{\mathrm{c}}$ & 1.0           & $\lambda_{\mathrm{d}}$ & 0.1             \\
Mixed precision        & \texttt{bf16} &                       &                  \\
\bottomrule
\end{tabular}
\vspace{-2mm}
\end{table}

\paragraph{VQ tokenizer architecture.}
LlamaGen-XL and Janus-Pro share a VQ-VAE tokenizer with $16{\times}$ spatial downsampling. %
The codebook has $|\mathcal{C}|{=}16{,}384$ entries, embedding dimension $d{=}8$, $\ell_2$-normalized before lookup. %
The encoder/decoder is a convolutional ResNet with channel multipliers $[1,1,2,2,4]$ and self-attention at $16{\times}16$. %
A $256{\times}256$ input yields a $256$-token raster-scan sequence. %

\paragraph{Reward models.}
We use two reward models: \textit{CLIP Score}~\citep{radford2021learning} (ViT-g/14 cosine similarity, scaled by $100$) and \textit{HPSv2}~\citep{wu2023hpsv2} (a human-preference scorer trained on large-scale pairwise judgments). %
Both are frozen; their gradients flow through the decoder in $\mathcal{L}_{\mathrm{reward}}$ but are detached before the non-differentiable sampling boundary. %

\paragraph{Compute footprint.}
RankE introduces a minimal training-time overhead compared to the standard frozen-decoder RL baseline. %
While holding the discriminator and EMA decoder in memory increases peak VRAM from 33~GB to 56~GB, the alternating update scheme itself adds no extra kernel-level or infrastructure cost. %
Specifically, a full training run of 6k steps for RankE takes approximately 20 hours on 8$\times$A100 GPUs, compared to 19 hours for the single-stage GRPO baseline. %
This represents a modest temporal overhead of only 5\%, which is strictly bounded and directly proportional to the additional parameters held in memory rather than the frequency of updates. %

\section{Training Algorithm}
\label{sec:algorithm_details}

Algorithm~\ref{alg:joint_training} summarizes the end-to-end training procedure. %
The framework alternates between policy optimization for the AR generator and decoder refinement under the multi-objective loss in Eq.~\ref{eq:decoder}. %

\algnewcommand{\Phase}[1]{\vspace{1.5mm}\State \textbf{\textcolor{teal!80!black}{\# #1}}}

\algnewcommand{\RightComment}[1]{\hfill \textcolor{gray}{\textit{$\triangleright$ #1}}}

\begin{algorithm}[t]
\caption{RankE: End-to-End Post-Training for Autoregressive Text-to-Image Generation}
\label{alg:joint_training}
\begin{algorithmic}[1]
\Require Text prompts $\mathcal{T}$, VQ-VAE encoder $E$ (frozen), decoder $D_\phi$, AR policy $\pi_\theta$, reference policy $\pi_{\mathrm{ref}}$
\Require Reward model $R$, discriminator $D_{\mathrm{disc}}$
\Require Hyperparameters: $\lambda_r,\lambda_g,\lambda_c,\lambda_d,\;\beta$
\Ensure Optimized policy $\pi_\theta^*$ and decoder $D_\phi^*$
\vspace{1mm}
\State \textbf{Initialize:}
  $\pi_{\mathrm{ref}} \leftarrow \pi_\theta$,\;
  $\pi_\theta^{\mathrm{EMA}} \leftarrow \pi_\theta$,\;
  $D_\phi^{\mathrm{EMA}} \leftarrow D_\phi$
\For{each batch $\{t_i, x_i^{\mathrm{gt}}\}_{i=1}^B \sim \mathcal{T}$}
    \Phase{Sampling \& Reward}
    \State Sample $\{z_i^k\}_{k=1}^G \sim \pi_\theta(\cdot \mid t_i)$
    \State $z_i^{\mathrm{gt}} \leftarrow E(x_i^{\mathrm{gt}})$;\;
           $\hat{x}_i^k \leftarrow D_\phi(z_i^k)$;\;
           $r_i^k \leftarrow R(t_i,\, \hat{x}_i^k),\;\forall k$
    \Phase{Stage 1: Policy Update}
    \State $A_i^k \leftarrow r_i^k - \frac{1}{G}\sum_j r_i^j$;\quad
           $\mathrm{KL}_i^k \leftarrow \log \pi_\theta(z_i^k \mid t_i) - \log \pi_{\mathrm{ref}}(z_i^k \mid t_i)$
    \State $\mathcal{L}_{\mathrm{GRPO}} \leftarrow
           -\frac{1}{BG}\sum_{i,k}\!\Big[
             A_i^k \log \pi_\theta(z_i^k \mid t_i)
             \;-\; \beta\,\mathrm{KL}_i^k
           \Big]$
    \Phase{Stage 2: Decoder Update}
    \State $z_i^{\mathrm{ema}} \sim \pi_\theta^{\mathrm{EMA}}(\cdot \mid t_i)$;\;
           $\hat{x}_i^{\mathrm{ema}} \leftarrow D_\phi(z_i^{\mathrm{ema}})$;\;
           $\hat{x}_i^{\mathrm{recon}} \leftarrow D_\phi(z_i^{\mathrm{gt}})$
    \State $\mathcal{L}_{\mathrm{dec}} \leftarrow
             \lambda_r\,\|\hat{x}_i^{\mathrm{recon}} - x_i^{\mathrm{gt}}\|_1
           \;-\; \lambda_g\,\mathbb{E}[\log D_{\mathrm{disc}}(\hat{x}_i^{\mathrm{ema}})]
           \;-\; \lambda_d\,R(t_i,\, \hat{x}_i^{\mathrm{ema}})$
    \State \hspace{1.6em}$+\; \lambda_c\,
           \|D_\phi(z_i^{\mathrm{ema}}) - D_\phi^{\mathrm{EMA}}(z_i^{\mathrm{ema}})\|_2^2$
    \Phase{Optimization \& EMA}
    \State $\theta \leftarrow \theta - \alpha_\theta\,\nabla_\theta \mathcal{L}_{\mathrm{GRPO}}$;\quad
           $\phi \leftarrow \phi - \alpha_\phi\,\nabla_\phi \mathcal{L}_{\mathrm{dec}}$;\quad
           $\phi_{\mathrm{disc}} \leftarrow \phi_{\mathrm{disc}} - \alpha_{\mathrm{disc}}\,\nabla_{\phi_{\mathrm{disc}}} \mathcal{L}_{\mathrm{GAN}}$
    \State $\pi_\theta^{\mathrm{EMA}}
             \leftarrow \mu_\theta\,\pi_\theta^{\mathrm{EMA}}
             + (1{-}\mu_\theta)\,\pi_\theta$;\quad
           $D_\phi^{\mathrm{EMA}}
             \leftarrow \mu_\phi\,D_\phi^{\mathrm{EMA}}
             + (1{-}\mu_\phi)\,D_\phi$
\EndFor
\State \Return $\pi_\theta^*,\; D_\phi^*$
\end{algorithmic}
\end{algorithm}

\section{Training Data and Curation}
\label{app:data}

\paragraph{Training data.}
We construct a $15$K training set from the publicly available BLIP3o-60k instruction-tuning dataset~\citep{chen2025blip3} (high-quality image--text pairs curated with GPT-4o across diverse scenes, objects, and human gestures). %
Two-stage curation adapts this pool for reward-based post-training. %
(1)~Long captions from synthetic sources (DALL-E~3~\citep{betker2023improving}, JourneyDB~\citep{sun2023journeydb}) often exceed the $77$-token CLIP context window, causing reward truncation and unstable gradients; we use Qwen2.5-Instruct~\citep{qwen2024qwen25} to compress each caption into concise visual tags ($\leq 50$ words), retaining key attributes (subject, style, lighting, color) while removing prose. %
(2)~Stratified sampling balances domain coverage and evaluation-relevant capabilities (Table~\ref{tab:data_composition}). %

\begin{table}[ht]
\centering
\small
\caption{Composition of the $15$K training set.}
\label{tab:data_composition}
\renewcommand{\arraystretch}{1.1}
\begin{tabular}{llcc}
\toprule
\textbf{Source} & \textbf{Domain} & \textbf{Count} & \textbf{Ratio} \\
\midrule
JourneyDB~\citep{sun2023journeydb}   & Aesthetics / artistic styles & 4{,}500 & 30\% \\
DALL-E~3~\citep{betker2023improving} & Complex attribute binding     & 3{,}000 & 20\% \\
MS-COCO~\citep{lin2014microsoft}     & Natural scenes                & 3{,}000 & 20\% \\
Human Gestures                       & Structural accuracy           & 3{,}000 & 20\% \\
GenEval~\citep{ghosh2023geneval}     & Compositional skills          & 750     & 5\%  \\
Typography                           & Text rendering                & 750     & 5\%  \\
\midrule
\textbf{Total} & & \textbf{15{,}000} & \textbf{100\%} \\
\bottomrule
\end{tabular}
\end{table}

\paragraph{Caption summarization pipeline.}
We apply Qwen2.5-Instruct (7B)~\citep{qwen2024qwen25} with the prompt:
\begin{quote}
\small\ttfamily
Summarize the following image caption into concise visual tags
($\leq$50 words). Preserve: subjects, styles, lighting, colors,
composition. Remove: narrative prose, subjective descriptions,
filler words.\\[3pt]
Caption: \{original\_caption\}\\[3pt]
Visual Tags:
\end{quote}
This compression preserves semantic content relevant for reward evaluation while keeping the full caption within CLIP's effective context window. %

\end{document}